%% file: main.tex
\documentclass[smallcondensed]{svjour3}     
\usepackage{times}  
\usepackage[hyphens]{url}  
\usepackage{graphicx} 
\usepackage{amsmath,amssymb,amsfonts}
\usepackage[noend]{algpseudocode}

\usepackage{algorithm}
\usepackage{multirow}
\newcommand{\STAB}[1]{\begin{tabular}{@{}c@{}}#1\end{tabular}}
\usepackage{adjustbox}
\usepackage{subcaption}

\newcommand{\ours}{$\mathsf{FABBOO}$}

\begin{document}
\title{Online Fairness-Aware Learning with Imbalanced Data Streams}

\author{
    Vasileios Iosifidis \and
    Wenbin Zhang \and
    Eirini Ntoutsi}
\institute{
    Iosifidis V., Ntoutsi E. \at  L3S Research Center, Leibniz University of Hannover, Germany\\
    \email{\{iosifidis, ntoutsi\}@L3S.de}
    \and
    Zhang W. \at Carnegie Mellon University, USA \\
    \email{wenbinzhang@cmu.edu}
    }


\date{Received: date / Accepted: date}

\maketitle

\maketitle

\begin{abstract}
Data-driven learning algorithms are employed in many online applications, in which data become available over time, like network monitoring, stock price prediction, job applications, etc. The underlying data distribution might evolve over time calling for model adaptation as new instances arrive and old instances become obsolete. In such dynamic environments, the so-called data streams, fairness-aware learning cannot be considered as a one-off requirement, but rather it should comprise a continual requirement over the stream.
Recent fairness-aware stream classifiers ignore the problem of class imbalance, which manifests in many real-life applications,
and  mitigate discrimination mainly because they  ``reject'' minority instances at large due to their inability to effectively learn all classes.

In this work, we propose \ours, an online fairness-aware approach that maintains a valid and fair classifier over the stream. \ours~is an online boosting approach that changes the training distribution in an online fashion by monitoring stream's class imbalance and tweaks its decision boundary to mitigate discriminatory outcomes over the stream. 
Experiments on 8 real-world and 1 synthetic datasets from different domains with varying class imbalance demonstrate the superiority of our method over state-of-the-art fairness-aware stream approaches with a range (relative) increase [11.2\%-14.2\%] in balanced accuracy, [22.6\%-31.8\%] in gmean, [42.5\%-49.6\%] in recall, [14.3\%-25.7\%] in kappa and [89.4\%-96.6\%] in statistical parity (fairness).


\keywords{data streams \and fairness-aware classification \and class imbalance }
\end{abstract}

\section{Introduction}
\label{sec:intro}
\input{introduction}

\section{Basic Concepts and Problem Definition}
\label{sec:prelim}
\input{preliminaries}

\section{Related Work}
\label{sec:related}
\input{related}

\section{Online Fairness- and Class Imbalance-aware Boosting}
\label{sec:approach}
\input{approach}

\section{Evaluation}
\label{sec:exp}
\input{experiments}

\section{Conclusion}
\label{sec:conclusion}
\input{conclusions}

 \bibliographystyle{spmpsci}      
\bibliography{bibliography}

\end{document}

%% file: introduction.tex
Data-driven learning algorithms have become a necessity nowadays for many applications that generate huge amounts of data.
Their performance in many tasks is comparable or has even surpassed human performance~\cite{grace2018will} and therefore,  for many processes, human decisions are substituted by algorithmic ones.
Such a replacement, however, has raised a lot of concerns~\cite{calders2013unbiased} regarding the fairness, accountability and transparency of such methods in domains of high societal impact such as risk assessment, recidivism, predictive policing, etc.
For example, the Google's \textit{AdFisher} online recommendation tool was found to show significantly more highly paid jobs to men than women~\cite{datta2015automated}. 

As a result of the ever-increasing interest in issues of fairness and responsibility of data-driven learning algorithms, a large body of work already exists in the domain of fairness-aware learning~\cite{hu2020fairnn,iosifidis2019fae,iosifidis2018dealing,iosifidis2019adafair,kamiran2012data,kamiran2018exploiting,krasanakis2018adaptive,ntoutsi2020bias}. 
Only a few recent works, however,  investigate the problem of fair learning in non-stationary environments~\cite{iosifidis2019fairness,wenbin2019fairness}. 
Nonetheless, these methods ignore an important learning challenge, namely that the majority of datasets suffer from class imbalance (c.f., Table~\ref{tbl:datasets}). Class imbalance refers to the disproportion among classes; for a binary classification setting this means that one class, called \textit{minority} class, has significantly fewer examples than another class, called \textit{majority} class. If the imbalance problem is not tackled, the learner mainly learns the majority class and strongly misclassifies/rejects the minority. Such methods might appear to be fair for certain fairness definitions~\cite{verma2018fairness} that rely on parity in the predictions. 
In reality though the low discrimination scores are just an artifact of the low prediction rates for the minority class. This observation has been made in~\cite{iosifidis2019fae,iosifidis2018dealing,iosifidis2019adafair} but for the static case. We observe the same issue for the streaming case and therefore, we propose an imbalance monitoring mechanism based on which we adapt the weighted training distribution. 
Moreover, in a stream environment the decisions do not only have a short-term effect, but rather they might incur long-term effects. In case of discrimination, this means that discriminatory model decisions affect not only the immediate outcomes, but they might also affect future outcomes. For example,~\cite{national2004measuring} shows small wage gaps between college-educated blacks and whites when they are first hired, but the pay gap increased over the years as a result of cumulative discrimination effects. To this end, we propose to define discrimination cumulatively over the stream rather than based only on recent outcomes. 

In this work, we focus on fairness-aware stream classification in the presence of skewed class distributions. Our approach, called online \underline{fa}irness and class im\underline{b}alance-aware \underline{boo}sting (\ours), monitors the class imbalance as well as the discriminatory behavior of the stream learner on the incoming stream and updates a boosting classifier~\cite{chen2012online} to tackle these issues concurrently. In contrast to existing works such as~\cite{iosifidis2019fairness} that focus on short-term mitigation by considering only the behavior of the model within a data chunk, \ours~is able to mitigate cumulative discriminatory outcomes by accounting for discrimination from the beginning of the stream up to the current timepoint. Our experiments verify that when treating for \textit{short-term} discriminatory outcomes, the \textit{cumulative} effects can be substantially higher over time and therefore, a cumulative approach is better. Furthermore, \ours~can accommodate various \emph{parity-based} notions of fairness, namely statistical parity, equal opportunity and predictive equality. Finally, \ours~is able to adapt to underlying changes in the data distribution --the so-called \textit{concept drifts}-- by employing Adaptive Hoeffding Trees~\cite{bifet2009adaptive} as weak learners, thus (\textit{blindly}) adapting to concept drifts.  

Our contributions are summarized as follows: i) we propose \ours, a fairness- and class imbalance-aware boosting approach that is able to tackle class imbalance as well as mitigate different parity-based discriminatory outcomes, ii) we introduce the notion of \textit{cumulative fairness} in streams, which accounts for cumulative discriminatory outcomes, iii) our experiments, in a variety of real-world and synthetic datasets, show that our approach outperforms existing fairness-aware approaches.

This work is an extension of our previous work~\cite{iosifidis2020fabboo}. The major changes include: i) modifying the distribution update part by also reducing the majority weights, ii) extending \ours~to facilitate another parity-based notion of fairness, namely predictive equality, iii) adding two real-world datasets to the experimental evaluation, 
iv) adding a recently published state-of-the-art imbalance-aware stream classifier~\cite{bernardo2020csmote} for comparison, iv) providing a detailed analysis w.r.t \ours's hyper-parameters selection.


%% file: preliminaries.tex
Let $X$ be a sequence of instances $x_1, x_2, \cdots,$ arriving over time at timepoints $t_1, t_2, \cdots$, where each instance 
$x \in \mathbb{R}^d$. 
Similarly, let $Y$ be a sequence of corresponding class labels, such that each instance in $x\in X$ has a corresponding class label in $y \in Y$. 
Without loss of generality, we assume a binary classification problem, i.e., $y=\{+1,-1\}$, and we denote by $y^+$ ($y^-$) the positive (negative, respectively) population segments.
We denote the classifier by $f(): X\rightarrow y$.
We follow the \emph{online learning setting}, where new instances from the stream are processed one by one. For each new instance $x$ arriving at $t$, its class label $f_{t-1}(x)$ is predicted by the current model $f_{t-1}()$. The true class label of the instance is revealed to the learner before the arrival of the next instance, and it is used for model updating, thus resulting into the updated model $f_t()$. This setup is known
as first-test-then-train or prequential evaluation~\cite{gama2010knowledge}. 
We assume that the underlying stream distribution is \emph{non-stationary}, that is, the characteristics of the stream might change with time leading to \emph{concept drifts}, i.e., changes in the joint distribution so that $P_{t_i}(X,y) \neq P_{t_j}(X,y)$ for two different timepoints $t_i$ and $t_j$. We are particularly interested in real concept drifts, that is when $P_{t_i}(y|x) \neq P_{t_j}(y|x)$, as such changes make the current classifier obsolete and call for model update.
Moreover, we consider the scenario where the stream population is \emph{imbalanced}, that is, one of the classes dominates the stream impacting the learning ability of the classifiers that traditionally tend to ignore the minority to foster generalization and avoid overfitting~\cite{weiss2004mining}. We do not require the minority class to be predefined and fixed over the stream. Instead, we assume that this role might alternate between the two classes.

We also assume the existence of a protected feature \textit{S}, e.g., gender or race, which is binary with values $S = \{z, \overline{z}\}$, e.g., gender=\{female, male\}; we refer to $z$, $\overline{z}$ as protected, non-protected group respectively~\footnote{S definition could also be extended to cover feature combinations, for example, race and gender}. 
Traditional \emph{fairness-aware classification} aims to learn a mapping $f(): X \rightarrow y$ that accurately maps instances $x$ to their correct classes without discriminating between the protected and non-protected groups. The discrimination is assessed in terms of some fairness measure. Formalizing fairness is a hard topic per se, and there has already been a lot of work in this direction. For example, \cite{verma2018fairness} overview more than twenty measures in the fairness-aware learning literature, each of which might be appropriate for different applications. 
In this work, we focus on \emph{parity-based} notions of fairness that compare model's behavior between the protected and non-protected groups and in particular, on  \textit{statistical parity}~\cite{kamiran2012data}, \textit{equal opportunity}~\cite{hardt2016equality} and \textit{predictive equality}~\cite{verma2018fairness}. 

 \emph{Statistical parity} (S.P.) measures the difference in the probability of a random individual drawn from the non-protected group $\overline{z}$ to be predicted as positive and the probability of a random individual drawn from the protected group $z$ to be predicted as positive:
\begin{equation}\footnotesize
    \label{eq:statistical_parity}
    S.P. = P(f(x)=y^+|\overline{z}) - P(f(x)=y^+|z)
\end{equation}\normalsize

S.P. values lie in the [-1, 1] range, with 0 meaning that the decision does not depend on the protected attribute (aka fair), 1 meaning that the protected group is totally discriminated (aka discrimination), and -1 that the non-protected group is discriminated (aka reverse discrimination).
S.P. does not take into account the real class labels, and therefore meeting the S.P. requirement might result into unqualified individuals being assigned to the positive class, thus causing \textit{reverse discrimination}. 

\emph{Equal opportunity} (EQ.OP.)~\cite{hardt2016equality} resolves this issue by measuring the difference in the True Positive Rates (TPR) between the two groups, i.e.,:
\begin{equation}\footnotesize 
    \label{eq:equal_opportunity}
    EQ.OP. = P(f(x)=y^+|\overline{z},y^+) - P(f(x)=y^+|z,y^+)
\end{equation}

Similar to Equal Opportunity, \emph{Predictive Equality} (P.EQ.)~\cite{verma2018fairness} measures the difference in the True Negative Rates (TNR) between the two groups, i.e.,:
\begin{equation}\footnotesize 
    \label{eq:pred_equality}
    P.EQ. = P(f(x)=y^-|\overline{z},y^-) - P(f(x)=y^-|z,y^-)
\end{equation}
Similarly to S.P., the values of EQ.OP and P.EQ also lie in the [-1, 1] range.

Our work investigates the problem of fair classification in a stream environment. \textit{Fairness-aware stream learning} refers to the problem of maintaining a valid and fair classifier over the stream. The term \textit{valid} refers to the ability of the model to adapt to the underlying population changes and deal with concept drifts. At the same time, the classifier should be fair according to the adopted fairness measure such as S.P., EQ.OP,. or P.EQ.
Ensuring fairness is much harder in such an online environment comparing to the traditional batch setting. First, the model should be continuously updated to reflect the underling non-stationary population. The typically accuracy-driven update of the model cannot ensure fairness, so even if the initial model was fair, its discriminatory behavior might get affected by the model updates. 
Second, small amounts of unfairness at each time point might accumulate into significant discrimination~\cite{national2004measuring} as the learner typically acts as an amplifier of whatever biases exist in the data and furthermore, reinforces its errors. 
Therefore, model updates should consider fairness as a permanent requirement over the stream and should take into account long term effects of discrimination.




%% file: related.tex
\textbf{Static Fairness-Aware Learning:}
Static or batch fairness-aware learning approaches have received a lot of attention over the recent years. Literature for bias mitigation in this area can be categorized into: i) pre-processing, ii) in-processing and iii) post-processing approaches.
Pre-processing approaches~\cite{calmon2017optimized,iosifidis2018dealing,kamiran2012data} focus on the data and aim to produce a ``balanced” dataset that
can then be fed into any learning algorithm. Different ``balancing'' techniques have been proposed from label swapping, known as massaging~\cite{kamiran2009classifying}, to increasing the representativeness of the protected group either via sampling~\cite{kamiran2012data} or via (semi-)synthetic data augmentation~\cite{iosifidis2018dealing} and transforming the feature space to remove attribute correlations with the protected attribute~\cite{calmon2017optimized}.  
In-processing approaches reformulate the classification problem by explicitly incorporating the model's discrimination
behavior in the objective function through regularization or constraints, or by training on latent target labels~\cite{iosifidis2019adafair,krasanakis2018adaptive}. Post-processing approaches~\cite{fish2016confidence,hardt2016equality,iosifidis2019fae} alter a model's predictions or adjust a model's decision boundary to reduce unfairness.

\noindent{\textbf{Stream Learning:}}
In stream learning, data arrive sequentially and their distributions can change over time, the so-called concept drifts~\cite{gama2014survey}. Concept drifts can be handled explicitly through \textit{informed adaptation}, where the model adapts only if a change has been detected, or implicitly through \textit{blind adaptation}, where the model is updated constantly to account for changes in the underlying data distributions. In addition, models developed for stream learning are categorized as \textit{incremental} and \textit{online}~\cite{wang2013learning}. Incremental models are trained in batches~\cite{forman2006tackling}, with the help of a chunk (window) of instances, while online models are updated continuously (per instance) to accommodate newly incoming examples~\cite{chen2012online}. 
The goal of stream classification is to maintain a valid classifier over the stream as new data arrive and old data become outdated. The maintenance depends on the underlying learning model, e.g., Naive Bayes classifiers~\cite{wagner2015ageing}, Decision Trees~\cite{bifet2009adaptive,matuszyk2013correcting}, KNN~\cite{losing2016knn}.

\noindent{\textbf{Stream Fairness-Aware Learning:}}
Stream fairness-aware learning approaches aim to mitigate discrimination in streaming environments where data arrive over time and concept drifts might occur which call for model adaptation~\cite{gama2010knowledge}. 
The problem has only recently attracted attention despite the common belief and evidence that bias and discrimination evolves across time, e.g., for gender bias~\cite{haines2016times}. In~\cite{iosifidis2019fairness}, the authors present a pre-processing chunk-based approach that removes discrimination in each chunk before updating the online classifier. In particular, the authors use label swapping/massaging to ensure a ``fair'' representation of protected and non-protected groups in each chunk, where fairness is measured via statistical parity on the predictions of the model. This approach however, tackles short-term discrimination effects, i.e., within each chunk. 
In~\cite{wenbin2019fairness}, an in-processing approach is proposed that incorporates the notion of statistical parity in the splitting criterion of a Hoeffding tree classifier, in order to select the best attribute for class separation and also for  fairness.


\noindent{\textbf{Stream class imbalance Learning:}}
Methods in this category aim to tackle the class imbalance problem while maintaining a model which adapts to concept drifts. As in batch-learning, if imbalance is not tackled the model will mainly learn the majority class. 
In~\cite{wang2013learning}, authors propose a framework which monitors stream's class imbalance ratio and use this information for re-sampling. In~\cite{wang2016online}, the authors convert classical class imbalance boosting methods such as AdaC2 and RUSBoost to online learners with the help of ADWIN change detector. A recently proposed method~\cite{bernardo2020csmote}, employs an online SMOTE strategy which re-samples the minority class from a sliding window and afterwards it updates an adaptive random forest ensemble equipped with ADWIN for change detection.  

In this paper we focus on fairness-aware learning for data streams with concept drifts and class imbalance; therefore, we select as competitors fairness-aware stream learners, namely~\cite{iosifidis2019fairness,wenbin2019fairness}.  
We also compare \ours~to the most recent state-of-the-art class imbalance stream learner~\cite{bernardo2020csmote}, although it does not tackle fairness.

%% file: approach.tex
\begin{figure*}[t!]
 \centering
 \includegraphics[width=1\textwidth]{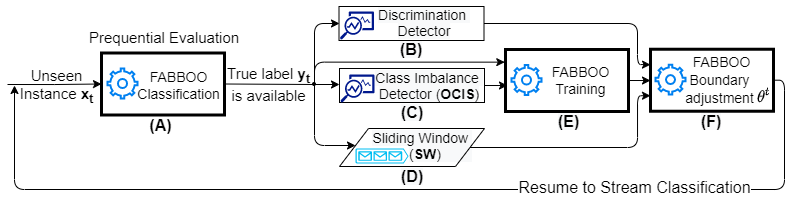}
 \caption{An overview of FABBOO}
 \label{fig:architecture}
\end{figure*}

An overview of \ours, standing for online \underline{fa}irness and class im\underline{b}alance-aware \underline{boo}sting, is shown in Figure~\ref{fig:architecture} (each component is highlighted with a letter in bold). In the online learning setting, the true labels become available after the classification of the unseen instances (\textbf{A}) (Sec.~\ref{sec:classification}). Our method consists of a \emph{class imbalance detector}  component (\textbf{C}) that monitors the class ratios over the stream (whole history) and adjusts the weights of the new training instances accordingly to ensure that the learner properly learns (\textbf{E}) both classes (Sec.~\ref{sec:class_monitoring}), while adapting to concept drifts via \textit{blind model adaptation}~\cite{gama2010knowledge}. 
In addition, the \textit{discrimination detector} component (\textbf{B}) (Sec.~\ref{sec:disc_monitoring}) monitors the cumulative discriminatory behavior of the learner and when it exceeds a user-defined tolerance threshold $\epsilon$, the decision boundary of the learner is adjusted (\emph{boundary adjustment} component (\textbf{F}), Sec.~\ref{sec:boundaryAdjustment}), with the support of a sliding window (\textbf{D}), to ensure that the learner is fair.

Our basic model is OSBoost~\cite{chen2012online} that generates smooth distributions over the training instances, and guarantees to achieve small error if the number of weak learners and training instances is large enough. OSBoost is an ensemble leaner $H$, which trains sequentially (homogeneous) weak learners, and comes with a set of predefined parameters: $\gamma \in [0,1]$ that is an online analog of the “edge” of the weak learning oracle, and $N \in \mathbb{Z}^+$ that is the number of online weak learners. We extend OSBoost to take into account class imbalance by changing the weighted instance distribution so that minority instances become more prominent during the training process (Sec.~\ref{sec:class_monitoring}) and tweak the model's decision boundary to account for fairness (Sec.~\ref{sec:disc_monitoring}). 

\subsection{Online Monitoring of Class Imbalance and Model Update}
\label{sec:class_monitoring}
In evolving data streams, the role of minority and majority classes can exchange and what is now considered to be minority might turn later into a majority class or vice versa~\cite{wang2013learning}. Knowing the class ratio over the stream is important for our method as it directly affects the instance weighting during training. Therefore, we keep track of the stream imbalance (whole history) using the online class imbalance monitor (OCIS) of~\cite{wang2013learning}.  
\begin{equation}\footnotesize 
    \label{eq:CCIS}
    OCIS_{t} = W_{t}^+ - W_{t}^- 
\end{equation}\normalsize
where $W_{t}^y$ is the percentage of class $y$ at timepoint $t$ maintained in an online fashion. In particular, upon the arrival of a new instance $x$ at timepoint $t$, the percentage of a class $y$ is updated as follows:
\begin{equation}\footnotesize
    \label{eq:class_monitor}
    W_{t}^y = \lambda\cdot W_{t-1}^y + (1-\lambda)\cdot\mathbb{I}[(y_{t},y)]
\end{equation}\normalsize
where $\mathbb{I}[(y_{t},y)]$ is an indicator function which equals to 1 if the true class label of $x_{t}$ is $y$, otherwise 0 and $\lambda \in [0,1)$ is a user-defined decay factor which controls the extent to which old class percentage information should be considered. The larger $\lambda$ is, the higher the contribution of historical information is, which however might hinder adaptation in case of concept drifts. A detailed analysis on the impact of $\lambda$ is shown in Sec.~\ref{sec:impact_of_L}. 

The imbalance index OCIS takes values in the $[-1,1]$ range, with 0 indicating a perfectly balanced stream and -1 or 1 indicating the total absence of one class. 

\noindent{\textbf{Model adaptation:}}
We extend OSBoost first for class imbalance; the pseudocode of the algorithm is shown in Algorithm~\ref{alg:train}. 
Upon the arrival of a new instance $x$ at timepoint $t$, the class imbalance status is updated (line 2) according to Equation~\ref{eq:CCIS}. 
Then, the weak learners are updated sequentially (lines 4-11) so that the predictions of model $H_i^t$ (line 6) affect the training of its successor model $H_{i+1}^t$ by changing the weight/contribution of instance $x$ to the model accordingly. 
The weight of instance $x$ is tuned per learner $H_i^t$ based on the error of the predecessor model $H_{i-1}^t$  on $x$, but also based on current class imbalance (lines 8-11). E.g., if at timepoint $t$ the $OCIS_{t} < 0$ and $x_t \in y^{+}$ then the $w_{i+1}$ will be increased. 
On the other hand, if at timepoint $t$ the $OCIS_{t} < 0$ and $x_t \in y^{-}$ then the $w_{i+1}$ will be decreased. Using this weighting strategy, \ours~increases/decreases the instances' weights according to the class imbalance monitor.

To summarize, traditional OSBoost performs error-based instance weight tuning but does not adjust for class imbalance. On the contrary, \ours~adjusts the instance weights also based on the dynamic class ratio (c.f.~Equation~\ref{eq:CCIS}) so that instances from both classes receive equal attention from the learners. Note that if the stream is balanced, i.e., $W_{t}^+ - W_{t}^-\approx 0$, the weights are only slightly affected. 

\begin{algorithm}[t!]\footnotesize 
\caption{FABBOO training procedure}
    \label{alg:train}
\begin{algorithmic}[1]
\Procedure{Train}{$x_{t},y_{t},\gamma, H^{t-1}_{1:N}$}      \Comment{$x_t$: newly arrived instance, $y_t$: label of $x_t$, $\gamma$: learning rate, $H^{t-1}_{1:N}$: current ensemble}
          \State $OCIS_{t} = W_{t}^+ - W_{t}^-$ \Comment{Update the class imbalance status}
        \State $w_1 = 1, q_0 = 0$ 
     \For{$i = 1$ to $N$}
        \State Train $H^{t}_i$ on $x_{t}$ with weight $w_i$  \Comment{Training of weak leaner $H_i$, where $i \in [1,N]$}
        \State $q_i = q_{i-1} + y_{t}\cdot H^{t}_i(x_{t}) - \frac{\gamma}{2+\gamma}$
        \State $w_{i+1} = min\{(1 - \gamma)^{q_{i}/2}, 1\}$  \Comment{OSBoost weight update}
       \If{$x_{t}\in y^+ $} \Comment{update $x_t$'s weight which belongs to$y^+$}
       \State $w_{i+1} = \frac{w_{i+1}}{1 + OCIS_{t}}$ 
        \EndIf
        \If{$x_{t}\in y^-$}  \Comment{update $x_t$'s weight which belongs to$y^-$}
      \State $w_{i+1} = \frac{w_{i+1}}{1 - OCIS_{t}}$
        \EndIf

      \EndFor
      \State \textbf{return} updated ensemble $H^t_{1:N}$
\EndProcedure
\end{algorithmic}
\end{algorithm}\normalsize 
 
\subsection{Online Monitoring of Cumulative Fairness and Boundary Adjustment}
\label{sec:disc_monitoring}
Methods which restore fairness only on \textit{short-term} (recent) outcomes fail to mitigate discrimination over time as discrimination scores that might be considered negligible when evaluated individually (i.e., at a single time point) might accumulate into significant discrimination in the long run~\cite{national2004measuring}. 
In this work, we aim to mitigate cumulative discrimination accumulated from the beginning of the stream in order to remove such long term discriminatory effects and adjust the decision boundary not only based on the recent behavior of the model, but also on its historical fairness-related performance.


Cumulative fairness monitoring accounts for discriminatory outcomes from the beginning of the stream until current time point $t$. 
We introduce the cumulative fairness notion for non-stationary environments. Cumulative fairness notions are updated per instance which makes them ideal for stream classification. In our work, we introduce cumulative fairness w.r.t. statistical parity~\cite{kamiran2012data}, equal opportunity~\cite{hardt2016equality} and predictive equality~\cite{verma2018fairness} as follows:

\begin{definition}{Cumulative Statistical Parity (Cum.S.P.)}
\label{eq:cumul_sp}\footnotesize 
    $$\frac{\sum\limits_{i=1}^{t} \mathbb{I}[f_i(x_i)=y^+|x_i\in\bar{z}]}{\sum\limits_{i=1}^{t} \mathbb{I}[x_i\in\bar{z}] + l} -  \frac{\sum\limits_{i=1}^{t} \mathbb{I}[f_i(x_i)=y^+|x_i\in z]}{\sum\limits_{i=1}^{t} \mathbb{I}[x_i\in z] + l}$$ 
\end{definition}\normalsize

\begin{definition}{Cumulative Equal Opportunity (Cum. EQ.OP.)}
\label{eq:cumul_eqop}\footnotesize
    $$\frac{\sum\limits_{i=1}^{t} \mathbb{I}[f_i(x_i)=y^+|x_i\in\bar{z},y^+_i]}{\sum\limits_{i=1}^{t} \mathbb{I}[x_i\in\bar{z}, y^+_i] + l} -  \frac{\sum\limits_{i=1}^{t} \mathbb{I}[f_i(x_i)=y^+|x_i\in z,y^+_i]}{\sum\limits_{i=1}^{t} \mathbb{I}[x_i\in z, y^+_i] + l}$$
\end{definition}\normalsize

\begin{definition}{Cumulative Predictive Equality (Cum. P.EQ.)}
\label{eq:cumul_pred_eq}\footnotesize
    $$\frac{\sum\limits_{i=1}^{t} \mathbb{I}[f_i(x_i)=y^-|x_i\in\bar{z},y^-_i]}{\sum\limits_{i=1}^{t} \mathbb{I}[x_i\in\bar{z}, y^-_i] + l} -  \frac{\sum\limits_{i=1}^{t} \mathbb{I}[f_i(x_i)=y^-|x_i\in z,y^-_i]}{\sum\limits_{i=1}^{t} \mathbb{I}[x_i\in z, y^-_i] + l}$$
\end{definition}\normalsize
\noindent where parameter $l$ is employed for correction in the early stages of the stream (0 division). 

Cum.S.P. Cum.EQ.OP or Cum.P.EQ. are maintained online using incremental counters updated with the arrival of new instances from the stream, and therefore, it is appropriate for stream applications where typically random access to historical stream instances is not possible. 
The cumulative fairness notions are employed by \ours~for discrimination monitoring. When their values exceed a user-defined discrimination tolerance threshold $\epsilon$, the decision boundary should be adjusted.

\noindent{\textbf{Decision boundary adjustment:}}
\label{sec:boundaryAdjustment}
Post-processing adjustment of the decision boundary for discrimination mitigation has been investigated in the literature, e.g.,~\cite{fish2016confidence,hardt2016equality}. 
Closer to our approach is~\cite{fish2016confidence}, where the authors adjust the decision boundary of an AdaBoost classifier based on the (sorted) confidence scores of misclassified instances of the protected group. However, in contrast to~\cite{fish2016confidence}, we deal with stream classification, and therefore, we do not have access to historical stream instances in order to adjust the boundary accurately. Except for the access-to-the-data constraint, another reason for not considering the whole history for the adjustment of the boundary is the non-stationary nature of the stream. In such a case, adjusting the boundary based on the whole history of the stream will hinder the ability of the model to adapt to the underlying data and will eventually hurt fair outcomes.

To overcome this issue, we use a sliding window model of a pre-defined size $M$ to approximate the optimal boundary tweak. In particular, we maintain a sliding window of size $M$ (see Figure~\ref{fig:architecture}) for each segment to allow for boundary adjustment for different parity-based notions based on each discriminated segment. 
In the case of statistical parity or equal opportunity, the only relevant sliding window is the one for the protected positive segment (denoted by $SW_z^+$). On the other hand, in the case of predictive equality, the relevant segment to be maintained is the negative segment (denoted by $SW_z^-$). 

For the boundary adjustment, there are two main steps: i) we approximate the optimal boundary adjustment by calculating how many instances need to be accepted/rejected in order to restore parity at timepoint $t$. ii) We proceed by shifting the boundary based on the confidence scores of the classified instances which reside in the sliding window. 
To estimate the number of examples $(n_t)$ which are needed in order to mitigate discrimination at timepoint $t$ for each fairness notion, we solve the cumulative fairness notions with respect to $(n_t)$. To calculate $n_t$ for cumulative statistical parity, we obtain:
\begin{equation}\footnotesize
\label{eq:position_sp} 
n_t = \Bigg\lfloor \sum_{i=1}^{t} \mathbb{I}[x_i\in z] \cdot \frac{\sum\limits_{i=1}^{t} \mathbb{I}[f_i(x_i)=y^+|x_i\in\bar{z}]}{ \sum\limits_{i=1}^{t} \mathbb{I}[x_i\in\bar{z}]} - \sum_{i=1}^{t} \mathbb{I}[f_i(x_i)=y^+|x_i\in z]\Bigg\rfloor
\end{equation}\normalsize

To estimate $n_t$ for equal opportunity, we follow the same rationale as previously: 
\begin{equation}\footnotesize
\label{eq:position_eqop} 
n_t = \Bigg\lfloor \sum\limits_{i=1}^{t} \mathbb{I}[x_i\in z,y_i^+] \cdot \frac{\sum\limits_{i=1}^{t} \mathbb{I}[f_i(x_i)=y^+|x_i\in\bar{z},y^+]}{ \sum\limits_{i=1}^{t} \mathbb{I}[x_i\in\bar{z}, y^+_i]} - \sum_{i=1}^{t} \mathbb{I}[f_i(x_i)=y^+|x_i\in z,y^+]\Bigg\rfloor
\end{equation}\normalsize

Predictive equality is identical to equal opportunity but for the negative class:
\begin{equation}\footnotesize
\label{eq:position_peq} 
n_t = \Bigg\lfloor \sum\limits_{i=1}^{t} \mathbb{I}[x_i\in z,y_i^-] \cdot \frac{\sum\limits_{i=1}^{t} \mathbb{I}[f_i(x_i)=y^-|x_i\in\bar{z},y^-]}{ \sum\limits_{i=1}^{t} \mathbb{I}[x_i\in\bar{z}, y^-_i]} - \sum_{i=1}^{t} \mathbb{I}[f_i(x_i)=y^-|x_i\in z,y^-]\Bigg\rfloor
\end{equation}\normalsize

In the offline case~\cite{fish2016confidence}, they tweak the boundary continuously until parity is achieved. In our work, we approximate the optimally fair decision boundary by considering as threshold the $n^{th}$'s instance confidence score. For this, the misclassified instances in $SW_z^+$ (or $SW_z^-$) are sorted based on the confidence scores in a descending order. The decision boundary is adjusted according to the $n_{t}$-th instance of the sorted window ($SW_z^+$ or $SW_z^-$). In particular, if $\theta^{t-1}$ is the decision boundary value (original value $\theta^{0}$ is 0.5) of the $n_{t-1}$-th, the fair-boundary is adjusted to $\theta^{t}$. Note that in the early stage of the stream, where the sliding window does not contain a sufficient number of instances, the boundary is tweaked based on the misclassified instance with the highest confidence within the window.

\subsection{\ours~Classification}
\label{sec:classification}
\ours~is an online ensemble of sequential weak learners that tackles class imbalance and cumulative discriminatory outcomes in the stream. Moreover, \ours~deals with concept drifts, through \textit{blind adaptation}, by employing a base learner that is able to react to concept drifts. In particular, we employ Adaptive Hoeffding Trees (\textbf{AHT})~\cite{bifet2009adaptive}  as weak learners; AHT is a decision-tree induction algorithm for streams that ensures DT model adaptation to the underlying data distribution by not only updating the tree with new instances from the stream, but also by replacing sub-trees when their performance decreases.  
 

The classification of a new unseen instance at time point $t$, i.e., $x_t$, is based on weighted majority voting and depends on its membership to $z$. If the instance does not belong to $z$ (i.e., it is a non-protected instance), then the standard boundary of the ensemble is used. Otherwise, the adjusted boundary is used. More formally (for Cum.S.P. and Cum.EQ.OP.):
\begin{equation}\footnotesize 
\label{eq:ensemble_disc}
f_t(x_{t}) =
 \begin{cases}
 y^+ & \quad  if~x_{t} \in z~and~H^t_{1:N}(x_{t}) \geq \theta^{t}\\
 H^t_{1:N}(x_{t}) & \quad \text{otherwise.}
 \end{cases}
\end{equation}\normalsize
where $N$ is the number of weak learners of the ensemble, and $\theta^t$ is the fair adjusted boundary at timepoint $t$. 
Similarly, for Cum.P.EQ. the decision of $f_t(x_{t})$ is $y^-$ is the confidence score is greater than the boundary at the time point $t$. Note that the adjustment of the boundary based on $\theta^t$ is applied at the ensemble level. 


%% file: experiments.tex
In this section, we introduce the employed competitors as well as variants of FABBOO\footnote{Data and source code are available at: \url{https://iosifidisvasileios.github.io/FABBOO}} (Sec.~\ref{sec:baselines}) that help us to demonstrate the behavior of FABBOO's individual components. The employed datasets as well as the performance measures are given in Sec.~\ref{sec:baselines}. Since Cum.P.EQ. is mirrored to Cum.EQ.OP. (only the target class changes from positive to negative), we have omitted the experiments since they show similar behavior as to Cum.EQ.OP.  
For the comparison to the competitors in Sec.s~\ref{sec:sp_results} and \ref{sec:eqop_results}, we set $\lambda= 0.9$, $M=2,000$. We also set $N=20$ for all the ensemble methods and a very small value $\epsilon=0.0001$, which means no tolerance to discriminatory outcomes. For the prequential evaluation of the non-stream datasets, we report on the average of 10 random shuffles (same as in~\cite{iosifidis2019fairness,wenbin2019fairness}), including the standard deviation. 
Furthermore, we provide a detailed analysis w.r.t: i) trade-off between predictive performance versus execution time in Sec.~\ref{sec:impact_of_t} (impact of $N$), ii) varying class imbalance overtime in Sec.~\ref{sec:impact_of_L} (impact of $\lambda$), and iii) concept drifts and boundary adjustment in Sec.~\ref{sec:impact_of_M} (impact of $M$).


\subsection{Competitors, Datasets and Metrics}
\label{sec:baselines}

\noindent{\textbf{Competitors}:} We evaluate FABBOO against two recent state-of-the-art fairness-aware stream classifiers~\cite{iosifidis2019fairness,wenbin2019fairness}, the fairness-agnostic non-stationary OSBoost~\cite{chen2012online} and imbalance-aware non-stationary and fairness agnostic CSMOTE~\cite{bernardo2020csmote}. We also employ two variations of FABBOO to show the impact of its different components, namely class imbalance and cumulative fairness. All methods employ AHTs~\cite{bifet2009adaptive} as weak learners and therefore are able to handle concept drifts. The only exception are FAHT~\cite{wenbin2019fairness} which is an incremental Hoeffding Tree that does not tackle concept drifts and CSMOTE~\cite{bernardo2020csmote} which is build on top of adaptive random forest equipped with ADWIN to handle concept drifts. An overview follows:

\begin{enumerate}

\item \textbf{Fairness Aware Hoeffding Tree (FAHT)~\cite{wenbin2019fairness}:} FAHT is an extension of the Hoeffding tree for fairness-aware learning that extends the typically employed information gain split attribute selection criterion to also include fairness gain (based on the statistical parity fairness measure). FAHT grows the tree by jointly considering information- and fairness-gain, however, it does not handle concept drifts nor class imbalance.


\item \textbf{Massaging (MS)~\cite{iosifidis2019fairness}:} a chunk-based model-agnostic stream fairness-aware learning approach which minimizes statistical parity on recent discriminatory outcomes. In particular, it detects and mitigates discrimination within the current chunk by performing label swaps, also known as ``massaging''~\cite{kamiran2009classifying} and retrains the model based on the ``corrected" chunk. MS is dealing with concept drifts by blind adaptation (using any adaptive learner), but considers only short-term discrimination outcomes, i.e., within the chunk, and does not account for class imbalance. We use default chunk size of 1,000 instances. 



\item \textbf{Online Smooth Boosting (OSBoost)~\cite{chen2012online}:} OSBoost does not consider fairness nor class imbalance.  

\item \textbf{Continuous SMOTE (CSMOTE)~\cite{bernardo2020csmote}:} CSMOTE does not consider fairness but it tackles class imbalance by re-sampling the minority class from a sliding window. We initialize CSMOTE with its default hyper-parameters. 

\item \textbf{Online Fair Imbalanced Boosting (OFIB):} A variation of FABBOO that does not account for class imbalance i.e., it does not use OCIS during training. This variation helps to show the importance of tackling class imbalance. 

\item \textbf{Chunk Fair Balanced Boosting (CFBB):} A variation of FABBOO that tackles short-term, instead of cumulative, discrimination. This variation helps to show the importance of long term fairness assessment. Instead of accounting for discrimination from the beginning of the stream, it monitors a chunk of 1,000 instances.
\end{enumerate}

\noindent{\textbf{Datasets}:}
\begin{table*}[t!]
\caption{An overview of the datasets.}
\centering
\begin{adjustbox}{width=1\textwidth,center}
\begin{tabular}{lccccccccc}
\hline
 & \#Instances & \#Attributes & Sen.Attr. & $z$ & $\bar{z}$ & Class ratio (+:-) & Stream & Positive class & Source
 \\ \hline
Adult Cen. & 45,175 & 14 & Gender & Female & Male & 1:3.03 & - & \textless{}50K & \cite{dua2017}\\
Bank & 40,004 & 16 & Marit. Status & Married & Single & 1:7.57 & - & subscription & \cite{dua2017}\\
Compas & 5,278 & 9 & Race & Black & White & 1:1.12 & - & recidivism & \cite{larson2016we} \\
Default & 30,000 & 24 & Gender & Female & Male & 1:3.52 & - & default payment & \cite{dua2017}\\
Kdd Cen. & 299,285 & 41 & Gender & Female & Male & 1:15.11 & - & \textless{}50K & \cite{dua2017}\\
Law Sc. & 18,692 & 12 & Gender & Female & Male & 1:9.18 & - & bar exam & \cite{wightman1998lsac}\\
Loan & 21,443 & 38 & Gender & Female & Male & 1:1.26 & \checkmark & paid &\cite{cortez2019}\\
NYPD & 311,367 & 16 & Gender & Female & Male & 1:3.68 & \checkmark & felony &\cite{birchall2015data}\\
synthetic & 150,236 & 6 & synth. & synth. & synth. & 1:3.13 & \checkmark & synth. &\cite{iosifidis2019fairness}\\
\hline
\end{tabular}
\end{adjustbox}
\label{tbl:datasets}
\end{table*}
To evaluate FABBOO, we employ a variety of real-world as well as synthetic datasets which are summarized in Table~\ref{tbl:datasets}. The datasets vary in terms of class imbalance, dimensionality and volume. 
Same as in~\cite{iosifidis2019fairness,wenbin2019fairness}, we use \textit{Adult census} dataset (Adult) and \textit{Kdd Census} dataset (Kdd Cen.) as well as \textit{Bank} dataset, \textit{Compas} dataset,
\textit{Default} dataset and \textit{Law School (Law Sc.)} dataset by randomly shuffling them, since they are not stream datasets. 
We also employ \textit{Loan}, \textit{NYPD} and a \textit{synthetic} dataset, all of which have temporal characteristics. 
For the synthetic dataset, we follow the authors' initialization process~\cite{iosifidis2019fairness}, where each attribute corresponds to a different Gaussian distribution, and moreover, we inject class imbalance and concept drifts to the stream. Concept drifts in this scenario are implemented by shifting the mean average of each Gaussian distribution (5 non-reoccurring concept drifts have been inserted at random points, see Figure~\ref{fig:overtime_sp} or~\ref{fig:overtime_eqop}). In addition, we have used further synthetic datasets to show the impact of \ours's hyper-parameters (Sec.~\ref{sec:impact_of_L} and~\ref{sec:impact_of_M}). We introduce each of them in the corresponding sections.

\noindent{\textbf{Metrics}:}
To evaluate the performance of FABBOO and competitors, we employ a set of measures which are able to show the performance in the presence of class imbalance. Same as in~\cite{ditzler2012incremental}, we employ \textit{balanced accuracy} (Bal.Acc.), \textit{Gmean}, \textit{Kappa}, and \textit{Recall}. For measuring discrimination, we report on \textit{cumulative statistical parity (Cum.S.P.)} in Sec.~\ref{sec:sp_results} and \textit{cumulative equal opportunity (Cum.EQ.OP.)} in Sec.~\ref{sec:eqop_results}.

\subsection{Results on cumulative statistical parity}
\label{sec:sp_results}
\input{statistical_results}

\subsection{Results on cumulative equal opportunity}
\label{sec:eqop_results}

\input{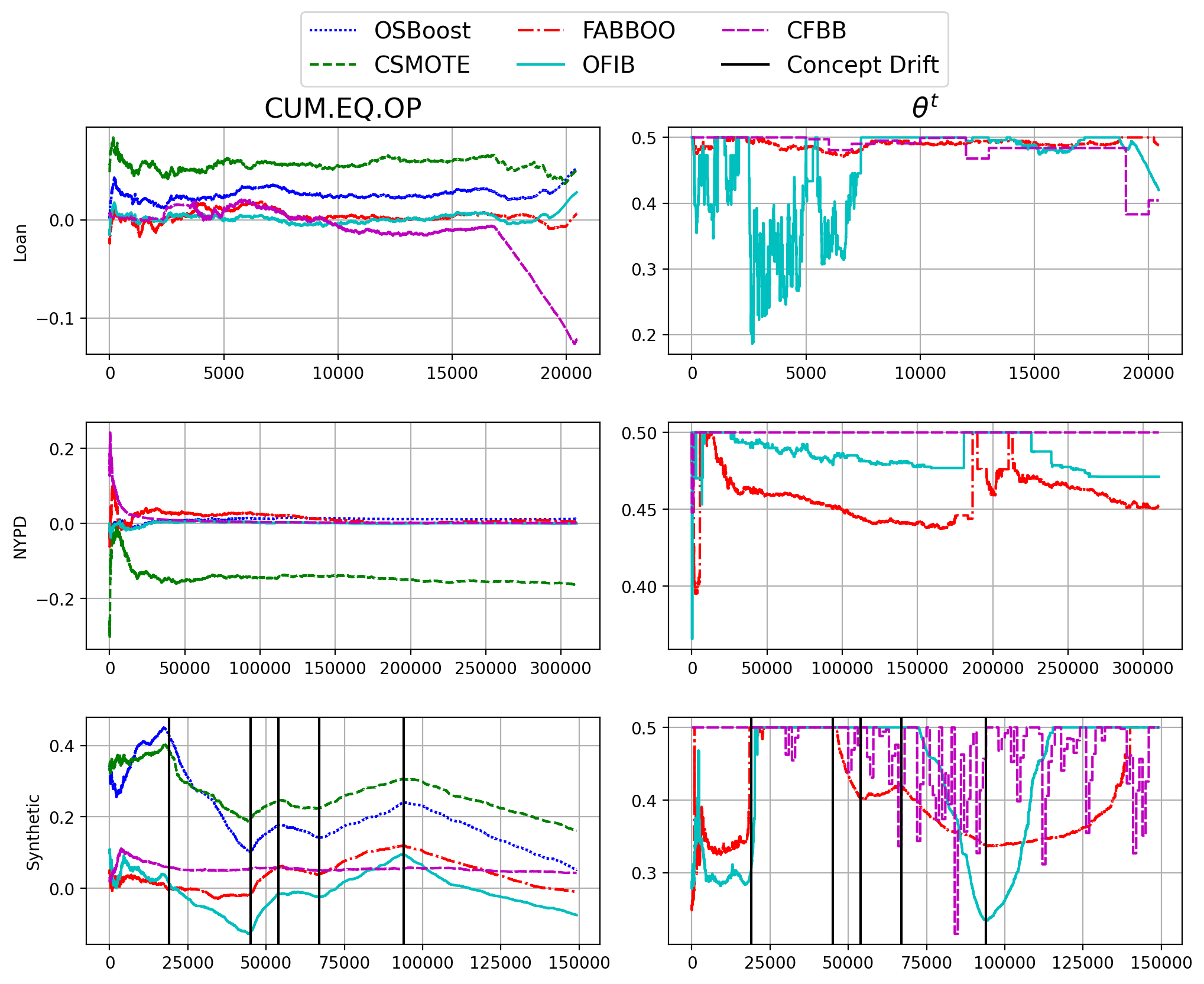}

\vspace{-.8mm}
\subsection{Performance vs execution time: Impact of $N$}
\label{sec:impact_of_t}
\input{impact_of_t}

\subsection{Varying class imbalance: Impact of $\lambda$}
\label{sec:impact_of_L}
\input{impact_of_L}

\subsection{Concept drifts and fair boundary adjustments: Impact of $M$}
\label{sec:impact_of_M}
\input{impact_of_M}

%% file: statistical_results.tex
\begin{table}[t!]
\centering
\caption{Overall predictive and fairness performance for Cum.S.P. (first and second best methods are in bold and underline, respectively)}
\label{tbl:results_sp}
\begin{adjustbox}{width=0.715\textwidth,center}
\begin{tabular}{llrrrrr} 
\hline
  & \multicolumn{1}{c}{Method} & \multicolumn{1}{c}{Bal. Acc. (\%)}  & \multicolumn{1}{c}{Gmean (\%)} & \multicolumn{1}{c}{Kappa (\%)}  & \multicolumn{1}{c}{Recall (\%)}  & \multicolumn{1}{c}{Cum.S.P.}  \\ 
\hline
\multirow{8}{*}{\STAB{\rotatebox[origin=c]{90}{Adult}}} 
& FAHT  & 73.08$\pm$0.5 & 70.1$\pm$0.8 & 51.02$\pm$0.8 & 52.45$\pm$1.4 & 0.1675$\pm$0.012\\ 
& MS  & 72.94$\pm$0.8 & 69.99$\pm$1.1 & 50.58$\pm$1.6 & 52.43$\pm$1.8 & 0.2366$\pm$0.011\\ 
& OSBoost  & 73.91$\pm$0.4 & 71.08$\pm$0.6 & \textbf{52.91$\pm$0.6} & 53.65$\pm$1 & 0.1783$\pm$0.006\\ 
& CSMOTE  & \textbf{78.12$\pm$0.7} & \textbf{78.08$\pm$0.8} & 47.91$\pm$2.1 & \underline{80.01$\pm$1.3} & 0.3237$\pm$0.02\\ 
& OFIB  & 74.13$\pm$0.2 & 72.71$\pm$0.3 & 48.17$\pm$0.3 & 59.7$\pm$0.7 & \underline{0.0019$\pm$0}\\ 
& CFBB  & 61.61$\pm$1.5 & 49.08$\pm$3.3 & 13.21$\pm$2 & \textbf{98.69$\pm$0.3} & 0.1697$\pm$0.022\\ 
& \ours~  & \underline{76.27$\pm$0.3} & \underline{76.04$\pm$0.4} & \underline{49.43$\pm$0.5} & 70.57$\pm$1.3 & \textbf{0.0018$\pm$0.002}\\ 
\hline
\multirow{8}{*}{\STAB{\rotatebox[origin=c]{90}{Bank}}}
& FAHT  & 62.51$\pm$1.6 & 51.84$\pm$3.1 & 32.17$\pm$3.1 & 27.73$\pm$3.5 & 0.0232$\pm$0.004\\ 
& MS  & 63.44$\pm$1.2 & 54.01$\pm$2.2 & 33.26$\pm$2.2 & 30.24$\pm$2.5 & 0.0835$\pm$0.006\\ 
& OSBoost  & 64.54$\pm$0.8 & 55.66$\pm$1.5 & 36.45$\pm$1.5 & 31.9$\pm$1.7 & 0.0322$\pm$0.002\\ 
& CSMOTE  & \underline{81.49$\pm$0.9} & \underline{81.48$\pm$0.9} & 40.88$\pm$1.9 & \underline{81.87$\pm$1.4} & 0.0772$\pm$0.005\\ 
& OFIB  & 67.94$\pm$0.8 & 61.88$\pm$1.3 & \underline{41.14$\pm$1.2} & 39.93$\pm$1.8 & 0.0037$\pm$0.001\\ 
& CFBB  & 70.24$\pm$1.3 & 67.07$\pm$2 & 15.52$\pm$1.6 & \textbf{90.97$\pm$0.9} & \underline{-0.0023$\pm$0.007}\\ 
& \ours~  & \textbf{81.82$\pm$0.5} & \textbf{81.65$\pm$0.5} & \textbf{48.01$\pm$0.5} & 76.66$\pm$1.6 & \textbf{0.0022$\pm$0.002}\\ 

\hline
\multirow{8}{*}{\STAB{\rotatebox[origin=c]{90}{Compas}}} 
& FAHT  & 64.33$\pm$0.7 & 63.99$\pm$0.9 & 28.8$\pm$1.5 & 58.26$\pm$3.1 & 0.232$\pm$0.019\\ 
& MS  & 64.61$\pm$0.9 & 64.45$\pm$1 & 29.25$\pm$1.7 & 61.48$\pm$3.8 & 0.5018$\pm$0.034\\ 
& OSBoost  & \underline{65.2$\pm$0.3} & \underline{64.85$\pm$0.4} & \textbf{30.57$\pm$0.7} & 58.52$\pm$1.1 & 0.273$\pm$0.019\\ 
& CSMOTE  & \textbf{65.29$\pm$0.3} & \textbf{65.23$\pm$0.3} & \underline{30.5$\pm$0.6} & \underline{65.18$\pm$2.8} & 0.2077$\pm$0.02\\ 
& OFIB  & 64.34$\pm$0.3 & 64.31$\pm$0.3 & 28.67$\pm$0.7 & 62.57$\pm$1.1 & 0.0195$\pm$0.015\\ 
& CFBB  & 55.62$\pm$2.8 & 39.8$\pm$14.4 & 10.84$\pm$5.6 & \textbf{89.88$\pm$9.4} & \textbf{0.0084$\pm$0.036}\\ 
& \ours~  & 64.33$\pm$0.4 & 64.32$\pm$0.4 & 28.58$\pm$0.8 & 64.49$\pm$0.8 & \underline{0.018$\pm$0.008}\\ 

\hline
\multirow{8}{*}{\STAB{\rotatebox[origin=c]{90}{Default}}} 
& FAHT  & 62.27$\pm$1.2 & 52.36$\pm$2.4 & 30.76$\pm$2.7 & 28.66$\pm$2.7 & 0.0164$\pm$0.003\\ 
& MS  & 63.38$\pm$0.5 & 54.62$\pm$0.9 & 33$\pm$1.3 & 31.26$\pm$1.1 & 0.1145$\pm$0.006\\ 
& OSBoost  & 63.34$\pm$0.8 & 54.52$\pm$1.8 & \underline{32.92$\pm$1.3} & 31.17$\pm$2.3 & 0.0195$\pm$0.003\\ 
& CSMOTE  & 59.28$\pm$0.9 & 55.19$\pm$1.9 & 10.89$\pm$1.6 & \underline{80.62$\pm$2.3} & 0.0202$\pm$0.004\\ 
& OFIB  & \underline{64$\pm$0.8} & \underline{56.06$\pm$1.9} & 31.85$\pm$1.3 & 33.21$\pm$2.5 & \underline{0.0027$\pm$0.002}\\ 
& CFBB  & 50.97$\pm$1.1 & 23.87$\pm$8.3 & 0.95$\pm$1.2 & \textbf{94.96$\pm$4} & -0.0241$\pm$0.126\\ 
& \ours~  & \textbf{67.49$\pm$0.5} & \textbf{66.31$\pm$0.7} & \textbf{33.96$\pm$2.1} & 55.37$\pm$3.1 & \textbf{0.0024$\pm$0.001}\\ 

\hline
\multirow{8}{*}{\STAB{\rotatebox[origin=c]{90}{Kdd Cen.}}}
& FAHT  & 63.28$\pm$3.2 & 51.78$\pm$6.5 & 35.98$\pm$6.4 & 27.52$\pm$6.7 & 0.03$\pm$0.009\\ 
& MS  & 61.29$\pm$1.2 & 48.11$\pm$2.7 & 32.07$\pm$2.6 & 23.42$\pm$2.6 & 0.0804$\pm$0.009\\ 
& OSBoost  & 64.77$\pm$0.2 & 54.94$\pm$0.3 & \underline{39.47$\pm$0.4} & 30.47$\pm$0.4 & 0.0377$\pm$0.002\\ 
& CSMOTE  & \underline{78.56$\pm$0.8} & \underline{78.38$\pm$0.9} & 30.15$\pm$0.3 & \underline{74.44$\pm$2.4} & 0.1156$\pm$0.003\\ 
& OFIB  & 66.64$\pm$0.2 & 59.12$\pm$0.3 & 37.56$\pm$0.5 & 35.9$\pm$0.4 & \underline{0.0007$\pm$0}\\ 
& CFBB  & 75.74$\pm$0.9 & 75.52$\pm$1 & 18.04$\pm$2.9 & \textbf{79.87$\pm$3.3} & 0.0334$\pm$0.012\\ 
& \ours~  & \textbf{79.04$\pm$0.2} & \textbf{78.83$\pm$0.3} & \textbf{39.94$\pm$0.1} & 64.35$\pm$0.5 & \textbf{0.0001$\pm$0.001}\\ 

\hline
\multirow{8}{*}{\STAB{\rotatebox[origin=c]{90}{Law Sc.}}} 
& FAHT  & 53.74$\pm$0.6 & 28.65$\pm$2.1 & 11.93$\pm$1.7 & 8.32$\pm$1.2 & 0.0088$\pm$0.003\\ 
& MS  & 56.71$\pm$1.4 & 38.29$\pm$4.2 & 19.29$\pm$3.3 & 15.11$\pm$3.4 & 0.0337$\pm$0.007\\ 
& OSBoost  & 57.19$\pm$0.6 & 39.41$\pm$1.5 & 21.06$\pm$1.4 & 15.77$\pm$1.2 & 0.0161$\pm$0.002\\ 
& CSMOTE  & \textbf{76.64$\pm$0.4} & \textbf{76.62$\pm$0.4} & \underline{28.28$\pm$1.2} & \underline{77.45$\pm$2.2} & 0.0216$\pm$0.004\\ 
& OFIB  & 58.33$\pm$0.7 & 42.54$\pm$1.6 & 23.44$\pm$1.5 & 18.46$\pm$1.4 & \underline{0.005$\pm$0.002}\\ 
& CFBB  & 67.75$\pm$2.2 & 66.79$\pm$3.3 & 13.74$\pm$2.6 & \textbf{77.88$\pm$2.8} & -0.0311$\pm$0.016\\ 
& \ours~  & \underline{74.12$\pm$0.7} & \underline{72.75$\pm$0.9} & \textbf{36.99$\pm$0.7} & 60.02$\pm$1.8 & \textbf{0.0018$\pm$0.002}\\ 

\hline
\multirow{8}{*}{\STAB{\rotatebox[origin=c]{90}{Loan}}} 
& FAHT  & 65.55& 65.54& 32.86& 66.52& \textbf{-0.0034}\\ 
& MS  & 66.83& 66.83& 33.41& 67.6& 0.5213\\ 
& OSBoost  & \underline{68.71}& \underline{68.18}& 37.82 & \underline{77.21}& 0.0556\\ 
& CSMOTE  & \textbf{69.17}& \textbf{68.94}& \textbf{38.51}& 74.77& 0.0358\\ 
& OFIB  & 68.68& 68.11& \underline{37.86} & \textbf{77.51} & 0.047\\ 
& CFBB  & 68.2& 60.34& 33.53& 36.41& -0.0644\\ 
& \ours~  & 66.85& 66.53& 33.91& 73.35& \underline{0.0317}\\ 

\hline
\multirow{8}{*}{\STAB{\rotatebox[origin=c]{90}{NYPD}}} 
& FAHT  & 50.11& 6.11& 0.35& 0.37& -0.0007\\ 
& MS  & \underline{56.95}& 41.02& \underline{18.38}& 17.44 & 0.0587\\ 
& OSBoost & 52.24&  24.33 & 6.45 & 6.01 & 0.0075\\ 
& CSMOTE & 56.22& \underline{54.56} &  11.01 & 42.68 & -0.1483\\ 
& OFIB & 52.23& 25.01& 6.75&  6.36& \underline{0.0003}\\ 
& CFBB & 50.05& 4.75& 0.05 & \textbf{99.88}& 0.0099\\ 
& \ours~ & \textbf{62.10}& \textbf{61.09}& \textbf{21.62}& \underline{50.95} &\textbf{0.0002}\\ 

\hline
\multirow{8}{*}{\STAB{\rotatebox[origin=c]{90}{synthetic}}} 
& FAHT  & 57.15& 42.5& 18.47& 18.95& 0.0832\\ 
& MS  & 62.43& 53.89& 29.97& 30.91& 0.1527\\ 
& OSBoost  & 63.43& 54.87& \underline{32.83}& 31.61& 0.0798\\ 
& CSMOTE  & \underline{67.84}& \underline{67.58}& 31.02& \underline{61.93}& 0.2613\\ 
& OFIB  & 63.76& 57.18& 31.78& 35.55& \underline{-0.0065}\\ 
& CFBB  & 57.71& 47.4& 8.71& \textbf{90.64}& 0.1232\\ 
& \ours~  & \textbf{68.3}& \textbf{67.53}& \textbf{33.72}& 58.03& \textbf{0.0017}\\ 

\hline
\end{tabular}
\end{adjustbox}
\end{table}

In this section, we compare our approach against the employed competitors for Cum.S.P., and report the overall results in Table~\ref{tbl:results_sp}. As we see, \ours~ is able to mitigate unfair outcomes and maintain the good predictive performance in terms of balanced accuracy, gmean, kappa, and recall for all datasets. 
More specifically, for \emph{Adult Cen.}, the best balanced accuracy is achieved by CSMOTE, which is designed to tackle class imbalance but not fairness, and \ours~ follows with 
only 1.85\%$\downarrow$ drop in balanced accuracy; However, the difference between these two models in terms of fairness is substantial (0.3219$\downarrow$ in Cum.S.P.). 
OFIB is also able to tackle discrimination but we observe a 2.4\%$\downarrow$ drop in balanced accuracy in contrast to \ours~. CFBB produces poor results in terms of predictive performance while the boundary adjustment is done per chunk; therefore it's delayed adjustment to the stream limits its ability to adapt effectively. This is also visible by the recall score. CFBB's delayed boundary adjustment allows more instances to be predicted positive; however, its overall predictive performance (balanced accuracy, gmean, kappa) is deteriorated. FAHT is able to lower discriminatory outcomes in contrast to OSBoost but \ours's predictive performance and fairness are significantly better than FAHT. Similar behavior can be observer by MS; however, MS is not even able to minimize discriminatory outcomes in contrast to OSBoost. Its chunk-based strategy limits its ability to tackle long term discrimination.

Overall, \ours~ achieves the best balanced accuracy, across all datasets, with an average score of 71.14\%, followed by CSMOTE with an average score of 70.17\%. In terms of discrimination, \ours~is the clear winner, across all datasets, with an average score of 0.0066, followed by OFIB with an average score of 0.0097. Although the difference in terms of discrimination is small, OFIB has an average balanced accuracy score of 64.45\%. CFBB achieves an average score of 0.0518 in terms of Cum.S.P, while FAHT and MS achieve an average score of 0.0628 and 0.1981, respectively. Fairness-aware competitors, FAHT and MS, are outperformed by \ours~in all reported metrics with a relative increase [11.2\%-14.2\%] in balanced accuracy, [22.6\%-31.8\%] in gmean, [42.5\%-49.6\%] in recall, [14.3\%-25.7\%] in kappa and [89.4\%-96.6\%] in Cum.S.P.

\begin{figure}[t!]
 \centering
 \includegraphics[width=.85\textwidth]{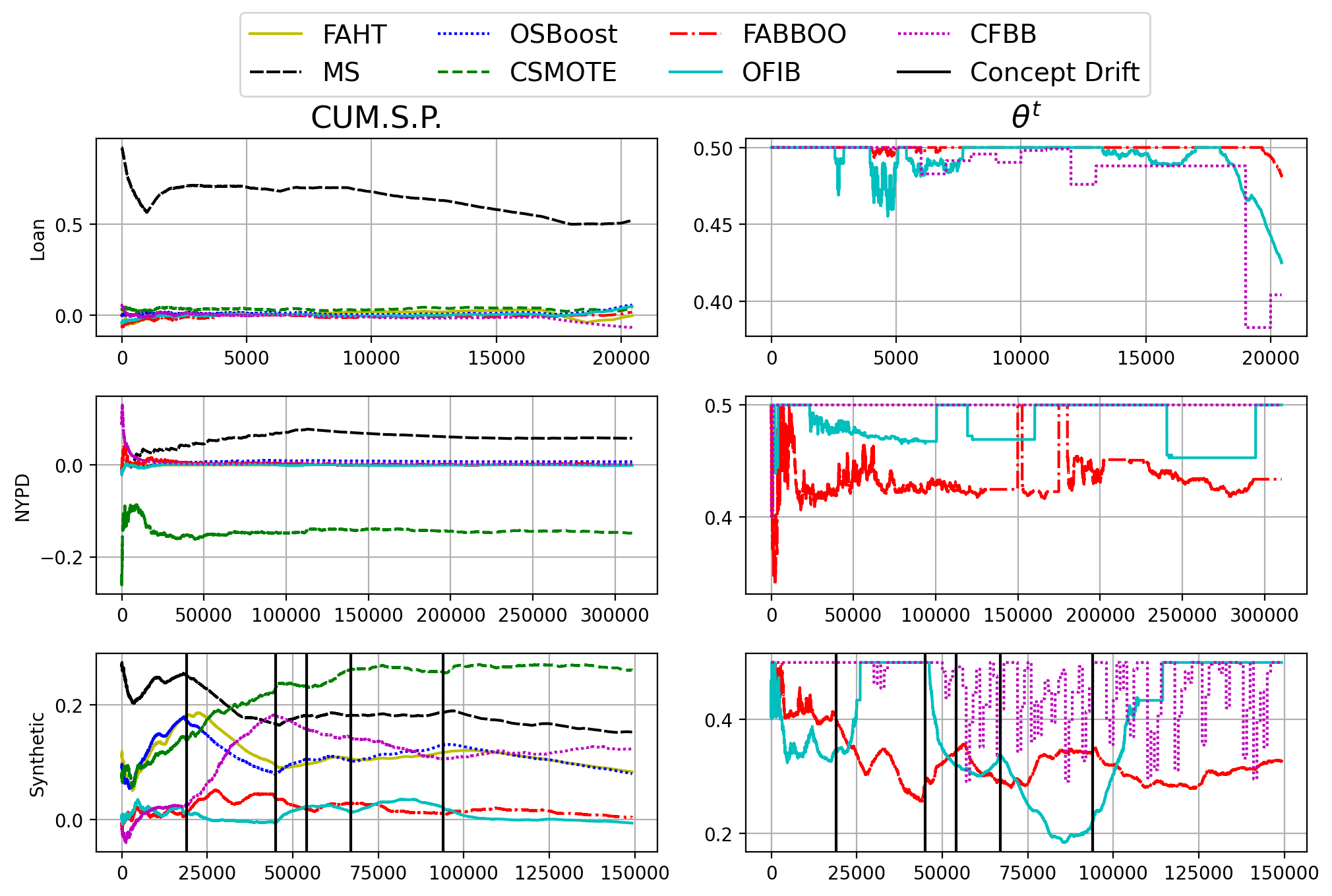}
 \caption{Cum.S.P. and boundary adjusting for Loan, NYPD and Synthetic datasets}
 \label{fig:overtime_sp}
\end{figure}

To get a closer look at the over time performance of the different methods, we show in Figure~\ref{fig:overtime_sp} the Cum. S.P. (left) and the required decision boundary adjustment (right), i.e., the boundary threshold $\theta^t$, for the datasets with temporal information (synthetic dataset contains concept drifts which are indicated by solid vertical lines). 
Looking at the Cum.S.P.(left), we see that for all datasets, CFBB is not able to mitigate discrimination except NYPD dataset for which it misclassifies almost all instances from majority class (c.f., Table~\ref{tbl:results_sp}), so the low unfairness is an artifact of the low prediction rates. MS falls in the same pitfall; by ``correcting'' the data based solely on the chuck it is not able to tackle unfair cumulative outcomes. Both CFBB and MS results show that a short-term consideration of fairness is unable to tackle discrimination propagation and reinforcement in the stream. 
The fairness-agnostic OSBoost and CSMOTE are also not able to tackle discrimination. The only exception for OSBoost is the \emph{NYPD} dataset. However, a closer look shows that low Cum.S.P. is only a result of vast rejecting the minority class (c.f., Table~\ref{tbl:results_sp}). On the other hand, \ours~and OFIB are able to tackle discrimination overtime, and outperform FAHT and MS. 

Looking at the required adjustments of the decision boundary (right), we notice that OFIB tends to produce higher boundary values than \ours. This is caused due to OFIB's inability to learn the minority class effectively; therefore, it rejects more minority instances from both protected and non-protected groups. For Loan dataset FABBOO and OFIB are performing similarly since the dataset is not severely imbalanced. Finally, we observe that CFBB has high fluctuation when adjusting the decision boundary due to its inability to adapt to underlying changes in data distributions w.r.t. fairness. Note that CFBB is tweaking slightly its decision boundary on NYPD dataset. 
 
\noindent\textbf{Conclusion:} \ours's predictive performance in contrast to imbalance-aware state-of-the-art competitor is better on average while its discriminatory outcomes are significantly lower than fairness-aware methods such as FAHT and MS. The results from \ours's variations, CFBB and OFIB, indicate the importance of the combination of class imbalance monitoring and boundary tweaking i.e., CFBB and OFIB are either unable to mitigate discrimination or maintain poor predictive performance. 

%% file: eqop_results.tex

For Cum.EQ.OP., we report the results of OSBoost, CSMOTE, OFIB, CFBB, and FABBOO on Table~\ref{tbl:results_eqop}. We exclude the other competitors such as FAHT and MS from these experiments since they are designed to mitigate unfair outcomes based on statistical parity. To the best of our knowledge, there are no fairness-aware stream methods that mitigate unfair outcomes based on equal opportunity. 
\begin{table}[t!]
\centering
\caption{Overall predictive and fairness performance for Cum.EQ.OP. (first and second best methods are in bold and underline, respectively)}
\label{tbl:results_eqop}
\begin{adjustbox}{width=.715\textwidth,center}
\begin{tabular}{llrrrrr} 
\hline
                  & \multicolumn{1}{c}{Method} & \multicolumn{1}{c}{Bal.Acc. (\%)} & \multicolumn{1}{c}{Gmean (\%)} & \multicolumn{1}{c}{Kappa (\%)}   & \multicolumn{1}{c}{Recall (\%)}  & \multicolumn{1}{c}{Cum.EQ.OP.}     \\ 
\hline
\multirow{5}{*}{\STAB{\rotatebox[origin=c]{90}{Adult}}} 
& OSBoost & 73.84$\pm$0.6 & 70.97$\pm$0.8 & 52.79$\pm$0.9 & 53.49$\pm$1.4 & 0.1906$\pm$0.022\\ 
& CSMOTE & \textbf{78.27$\pm$0.7} & \textbf{78.21$\pm$0.7} & 47.8$\pm$2 & \underline{80.91$\pm$1.3} & 0.1372$\pm$0.024\\ 
& OFIB & 74.7$\pm$0.5 & 72.32$\pm$0.7 & \underline{53.66$\pm$0.8} & 56.04$\pm$1.2 & \underline{0.0193$\pm$0.011}\\ 
& CFBB & 62.67$\pm$1.3 & 51.36$\pm$2.7 & 14.62$\pm$1.8 & \textbf{98.46$\pm$0.7} & 0.0252$\pm$0.008\\ 
& FABBOO & \underline{77.72$\pm$0.5} & \underline{76.91$\pm$0.7} & \textbf{55.35$\pm$0.5} & 66.61$\pm$1.8 & \textbf{0.0115$\pm$0.015}\\ 
\hline
\multirow{5}{*}{\STAB{\rotatebox[origin=c]{90}{Bank}}} 
& OSBoost & 65.29$\pm$0.7 & 57.15$\pm$1.3 & 37.38$\pm$1 & 33.77$\pm$1.7 & 0.0618$\pm$0.01\\ 
& CSMOTE & \underline{80.48$\pm$1.5} & \textbf{80.46$\pm$1.6} & 38.91$\pm$2.3 & \underline{80.78$\pm$2.9} & 0.0181$\pm$0.013\\ 
& OFIB & 66.6$\pm$0.8 & 59.55$\pm$1.5 & \underline{39.26$\pm$0.9} & 36.82$\pm$2 & \underline{0.0101$\pm$0.006}\\ 
& CFBB & 69.98$\pm$1.4 & 67.03$\pm$2.1 & 15.46$\pm$1.7 & \textbf{89.9$\pm$0.8} & -0.0235$\pm$0.004\\ 
& FABBOO & \textbf{80.53$\pm$0.4} & \underline{80.01$\pm$0.5} & \textbf{50.16$\pm$0.3} & 71.43$\pm$1.3 & \textbf{0.002$\pm$0.006}\\ 
\hline
\multirow{5}{*}{\STAB{\rotatebox[origin=c]{90}{ Compas}}} 
& OSBoost & \textbf{65.26$\pm$0.4} & \underline{64.93$\pm$0.4} & \textbf{30.67$\pm$0.8} & 58.9$\pm$1.5 & 0.2788$\pm$0.019\\ 
& CSMOTE & \underline{65.15$\pm$0.4} & \textbf{65.12$\pm$0.4} & \underline{30.21$\pm$0.9} & \underline{65.58$\pm$1.8} & 0.2012$\pm$0.042\\ 
& OFIB & 64.53$\pm$0.2 & 64.48$\pm$0.2 & 29.06$\pm$0.4 & 62.55$\pm$1.7 & \underline{0.0379$\pm$0.023}\\ 
& CFBB & 55.89$\pm$4.2 & 38.34$\pm$18.1 & 11.48$\pm$8.4 & \textbf{87.34$\pm$15.1} & 0.0466$\pm$0.032\\ 
& FABBOO & 64.55$\pm$0.3 & 64.54$\pm$0.3 & 29.03$\pm$0.5 & 64.53$\pm$1.3 & \textbf{0.0359$\pm$0.024}\\ 
\hline
\multirow{5}{*}{\STAB{\rotatebox[origin=c]{90}{Default}}}
& OSBoost & 63.16$\pm$0.3 & 54.22$\pm$0.8 & 32.56$\pm$0.6 & 30.78$\pm$1.1 & 0.0127$\pm$0.008\\ 
& CSMOTE & 59.07$\pm$0.4 & \underline{54.75$\pm$0.7} & 10.5$\pm$0.5 & \underline{81.21$\pm$0.9} & 0.0149$\pm$0.008\\ 
& OFIB & \underline{63.36$\pm$0.3} & 54.68$\pm$0.8 & \underline{32.77$\pm$0.5} & 31.37$\pm$1.1 & \textbf{0.0023$\pm$0.005}\\ 
& CFBB & 50.65$\pm$0.2 & 20.06$\pm$4.9 & 0.59$\pm$0.2 & \textbf{96.86$\pm$1.8} & 0.0144$\pm$0.008\\ 
& FABBOO & \textbf{67.65$\pm$0.5} & \textbf{66.3$\pm$0.9} & \underline{32.84$\pm$1.2} & 54.52$\pm$3.1 & \underline{0.006$\pm$0.004}\\ 

\hline
\multirow{5}{*}{\STAB{\rotatebox[origin=c]{90}{ Kdd Cen.}}} 
& OSBoost & 65$\pm$0.5 & 55.32$\pm$0.9 & 40.61$\pm$1.1 & 30.88$\pm$1 & 0.1512$\pm$0.013\\ 
& CSMOTE & \textbf{80$\pm$0.1} & \textbf{79.89$\pm$0.1} & 29.94$\pm$0 & \underline{75.82$\pm$0.4} & 0.084$\pm$0.003\\ 
& OFIB & 66.27$\pm$0.6 & 57.85$\pm$1 & \underline{41.11$\pm$1.2} & 33.95$\pm$1.2 & \underline{0.0075$\pm$0.004}\\ 
& CFBB & 74.9$\pm$0.5 & 74.73$\pm$0.5 & 16.46$\pm$0.7 & \textbf{79.89$\pm$1.2} & 0.0119$\pm$0.007\\ 
& FABBOO & \underline{78.32$\pm$0.2} & \underline{76.68$\pm$0.3} & \textbf{46.06$\pm$0.7} & 62.38$\pm$0.3 & \textbf{0.0034$\pm$0.002}\\ 
 
\hline
\multirow{5}{*}{\STAB{\rotatebox[origin=c]{90}{Law Sc.}}} 
& OSBoost & 57.47$\pm$0.6 & 40.2$\pm$1.6 & 21.69$\pm$1.4 & 16.43$\pm$1.3 & 0.0585$\pm$0.011\\ 
& CSMOTE & \textbf{76.8$\pm$0.4} & \textbf{76.78$\pm$0.4} & \underline{28.36$\pm$1.4} & \underline{77.86$\pm$1.6} & 0.0262$\pm$0.017\\ 
& OFIB & 58.17$\pm$0.8 & 42.12$\pm$2.1 & 23.04$\pm$1.7 & 18.11$\pm$1.8 & \underline{0.0256$\pm$0.011}\\ 
& CFBB & 67.57$\pm$1.6 & 66.41$\pm$2.3 & 13.11$\pm$2 & \textbf{79.3$\pm$3.2} & -0.0298$\pm$0.014\\ 
& FABBOO & \underline{73.88$\pm$0.5} & \underline{72.36$\pm$0.6} & \textbf{37.34$\pm$0.5} & 59.04$\pm$1.3 & \textbf{0.0132$\pm$0.01}\\ 

\hline
\multirow{5}{*}{\STAB{\rotatebox[origin=c]{90}{Loan}}} 
& OSBoost & \underline{68.71}& \underline{68.18}& \underline{37.86}& \underline{77.21}& 0.0525\\ 
& CSMOTE & \textbf{69.17}& \textbf{68.94}& \textbf{38.51}& 74.77& 0.0507\\ 
& OFIB & 68.65& 68.01& 37.82& \textbf{78.06}& \underline{0.0301}\\ 
& CFBB & 68.27& 60.45& 33.66& 36.54& -0.1219\\ 
& FABBOO & 66.78& 66.4& 33.82& 73.91& \textbf{0.0154}\\ 

\hline
\multirow{5}{*}{\STAB{\rotatebox[origin=c]{90}{NYPD}}} 
& OSBoost & 52.24&  24.33 & 6.45 & 6.01 & 0.0125\\ 
& CSMOTE & \underline{56.22}& \underline{54.56} & \underline{11.01}& 42.68 & -0.163\\ & OFIB & 52.23& 24.8& 6.66&  6.25& \textbf{0.0003}\\ 
& CFBB & 50.05& 4.78& 0.04 & \textbf{99.87}& 0.0115\\ 
& FABBOO & \textbf{61.98}& \textbf{60.79}& \textbf{22.59}& \underline{50.01} &\underline{0.0042}\\ 

\hline
\multirow{5}{*}{\STAB{\rotatebox[origin=c]{90}{synthetic}}} 
& OSBoost & 63.43& 54.87& \underline{32.83}& 31.61& 0.0518\\ 
& CSMOTE & \underline{67.84}& \underline{67.58}& 31.02& \underline{61.93}& 0.1619\\ 
& OFIB & 63.7& 56.81& 31.92& 34.88& -0.0732\\ 
& CFBB & 57.95& 47.96& 9.02& \textbf{90.47}& \underline{0.0428}\\ 
& FABBOO & \textbf{68.83}& \textbf{67.59}& \textbf{36.12}& 55.8& \textbf{-0.0013}\\ 

\hline
\end{tabular}
\end{adjustbox}
\end{table}
The results indicate that FABBOO performs good in terms of balanced accuracy, gmean, kappa, and recall in all datasets except Compass and Loan, which are balanced datasets (c.f., Table~\ref{tbl:datasets}). More concretely, for Adult Cen. dataset, the best balanced accuracy is achieved by CSMOTE followed by FABBOO (0.55\%$\downarrow$), the best Gmean is achieved by CSMOTE followed by FABBOO (1.3\%$\downarrow$), the best Kappa is achieved by FABOO followed by OFIB (1.6\%$\downarrow$), and the best recall is achieved by CFBB followed by CSMOTE (17.5\%$\downarrow$). FABBOO achieves the best Cum.EQ.OP. followed by OFIB (0.0078$\downarrow$), however OFIB rejects more instances from the positive class. Similar behavior can be observed in all datasets, where FABBOO is able to tackle class imbalance and mitigate unfair outcomes better than the other methods. OSBoost fails to learn the positive (minority) class, thus under-performs in almost all datasets. In some cases, it produces low discriminatory outcomes; however, this is a result of misclassifying huge portions of the minority class. 

Overall, \ours~achieves the best predictive performance with an average score of 71.13\% across all datasets, followed by CSMOTE with an average score of 70.3\%. In addition, FABBOO achieves the fairest results with an average score of 0.0103 followed by OFIB with an average score of 0.0229.

\begin{figure}[t!]
 \centering
 \includegraphics[width=.85\textwidth]{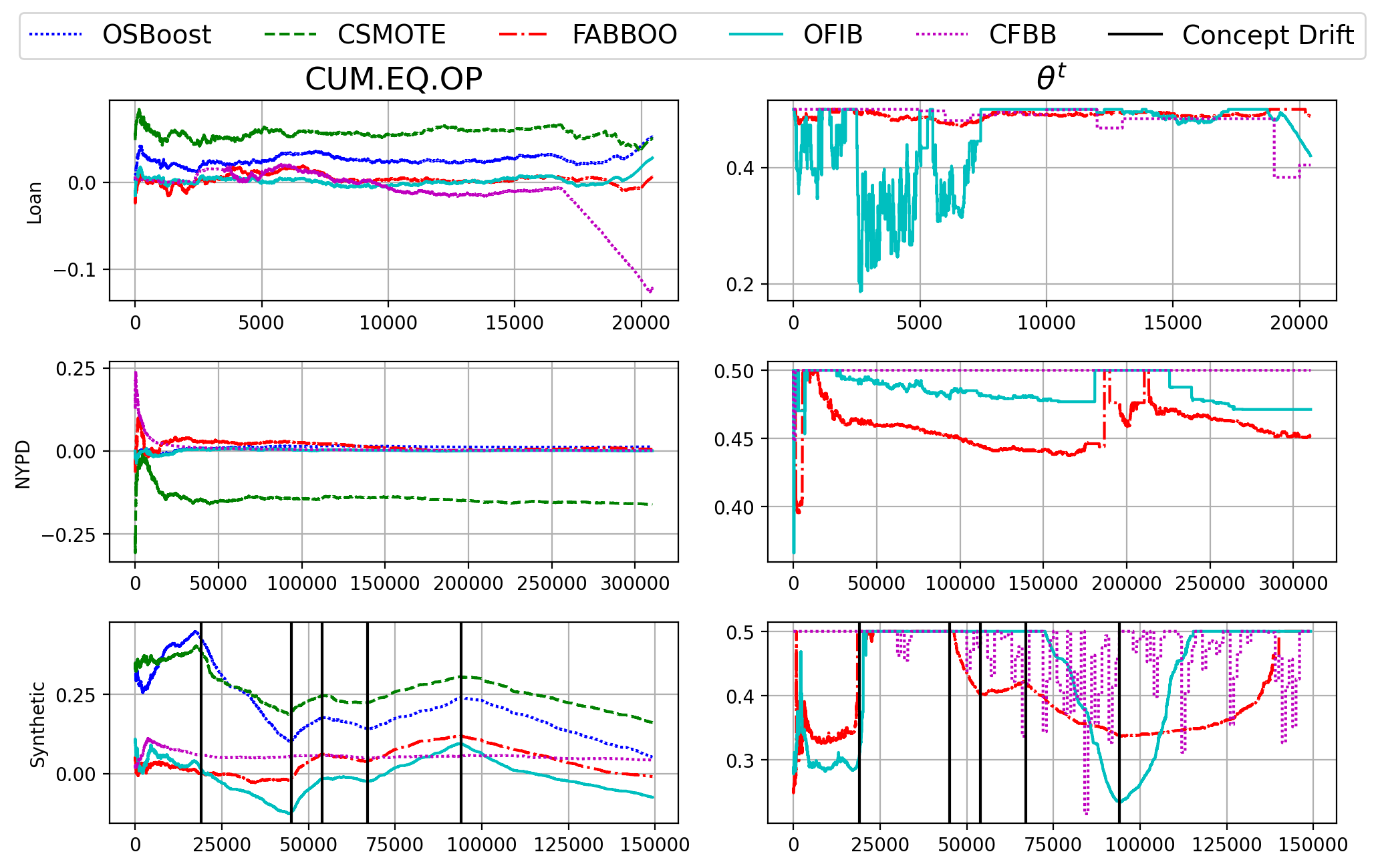}
 \caption{Cum.EQ.OP. and boundary adjusting for Loan, NYPD and Synthetic datasets}
 \label{fig:overtime_eqop}
\end{figure}

We also demonstrate how Cum.EQ.OP. and the decision boundary (FABBOO, OFIB and CFBB) vary over time for the stream datasets in Figure~\ref{fig:overtime_eqop}. Same as in Cum.S.P., we observe that \ours~is able to maintain low discriminatory outcomes over the stream. CFBB and OFIB achieve fairness only in the NYPD dataset by rejecting either the whole majority or minority class, respectively, and fail to mitigate unfairness on the other two datasets. Finally, CSMOTE is amplifying unfairness (and causes reverse discrimination in NYPD dataset) by re-sampling the instances. A reason for such behavior can be the amplification of existing encoded biases in the data through instance re-sampling (also reported in~\cite{iosifidis2019fae}).

\noindent\textbf{Conclusion:} Same as in the previous section, we observe that \ours~is able to maintain better predictive performance in contrast to CSMOTE. Although there is no competitor w.r.t equal opportunity, we showed that \ours~is able to mitigate discriminatory outcomes w.r.t equal opportunity. Similar behavior can also be observed for predictive equality. 

%% file: impact_of_t.tex
\begin{figure}
   \centering
   \begin{subfigure}[b]{0.5\textwidth}
 \includegraphics[width=\textwidth]{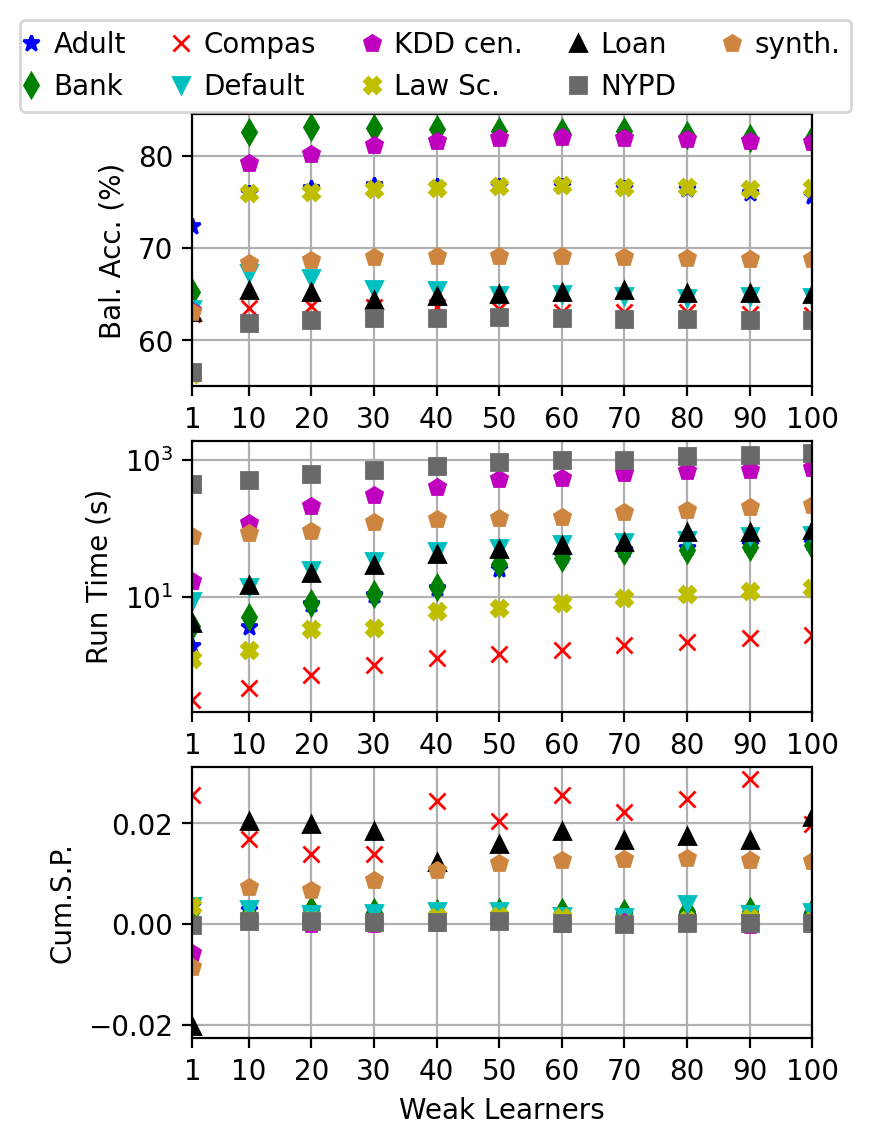}
        \caption{Results on Cum.S.P.}
       \label{fig:st_par_impact_of_N}
   \end{subfigure}\hfill
   \begin{subfigure}[b]{0.5\textwidth}
 \includegraphics[width=\textwidth]{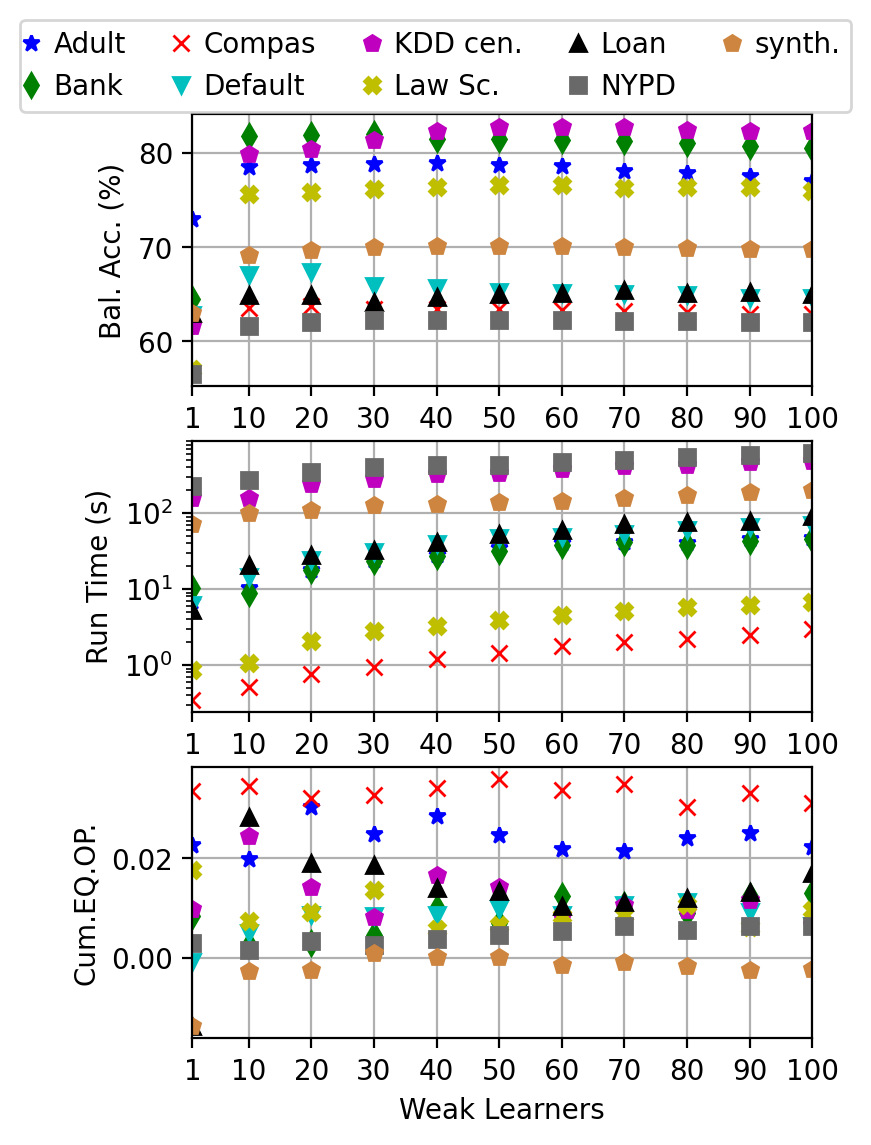}
              \caption{Results on EQ.OP.}
       \label{fig:table2}
    \end{subfigure}
    \caption{\ours: Impact of the number of weak learners $N$ on predictive and fairness performance for all datasets}\label{fig:eq_op_impact_of_N}
     \label{fig:impact_of_N}
\end{figure}

For analyzing the impact of $N$ (\#weak learners), we increase $N \in [1, 10, 20,...,100]$ and report on the average of 10 random shuffles (same as before) for each value of $N$. We show how much the predictive performance (balanced accuracy), fairness (Cum.S.P. or Cum.EQ.OP.) and execution time are affected by $N$ in Figure~\ref{fig:impact_of_N}.

In Figure~\ref{fig:st_par_impact_of_N}, \ours~is tuned to mitigate unfair outcomes based on statistical parity. We observe that for $N=1$ \ours~has poor predictive performance across all datasets; however, its unfair outcomes remain low. As $N$ increases, so does balanced accuracy but also the run time (linearly) e.g., for $N=1$ to $N=10$ balanced accuracy is increased by $5\%\uparrow$ for Adult cen., $18\%\uparrow$ for Bank, $2\%\uparrow$ for Compas, $4\%\uparrow$ for Default, $16\%\uparrow$ for KDD cen., $20\%\uparrow$ for Law Sc., $2\%\uparrow$ for Loan, $5\%\uparrow$ for NYPD and $5\%\uparrow$ for synthetic. Similar behavior can also be observed for equal opportunity (Figure~\ref{fig:eq_op_impact_of_N}). In terms of execution time, the addition of more weak learners increases the time linearly. Interestingly after a sufficient number of weak learners e.g., $N > 10$ \ours's results do not change significantly. 

\noindent\textbf{Conclusion:} From our experiments, we have observed that \ours's ability to maintain good predictive performance does not rely strongly on the hyper-parameter $N$ as long as $N$ is sufficiently large. As seen from the results, the performance does not change significantly, neither does \ours~'s ability to mitigate unfair outcomes. 

%% file: impact_of_L.tex
\begin{figure}[tp!]
 \centering
 \includegraphics[width=.75\textwidth]{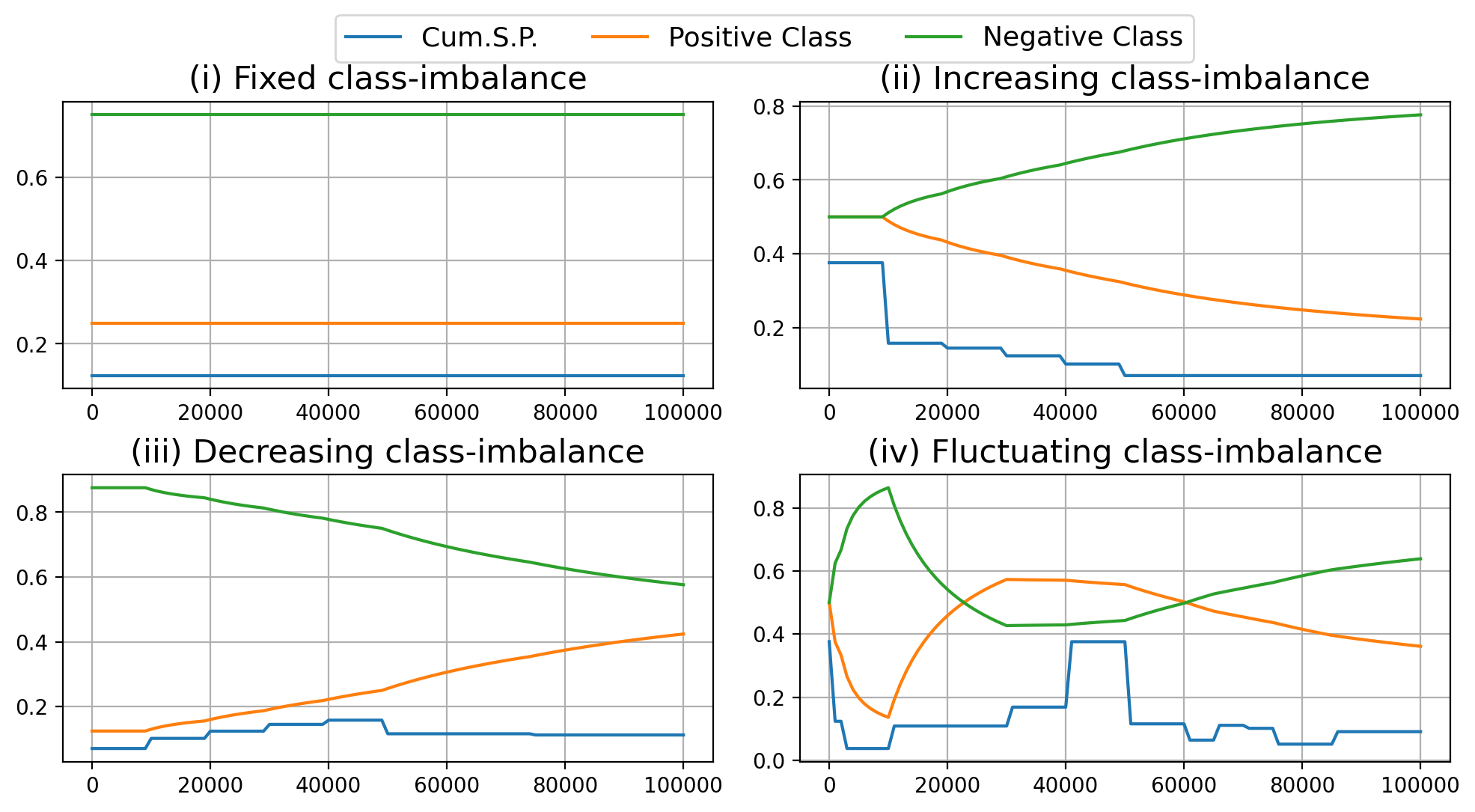}
 \caption{Synthetic datasets of varying class imbalance (employed for $\lambda$ evaluation)}
 \label{fig:synth_datasets_lamda}
\end{figure}

In this section, we analyze \ours's behaviour by varying the class imbalance ratio over time - recall that we do not assume a fixed minority class. We show how \ours~ is affected and the impact of parameter $\lambda$ on the online class imbalance monitor of Equation~\ref{eq:class_monitor}. For this purpose, we have generated synthetic data streams of varying class ratios over time (Figure~\ref{fig:synth_datasets_lamda}). For $\lambda$, we consider values in range of $[0, 0.99]$ with a $0.1$ incremental step. Recall that low $\lambda$ means lower contribution of historical data (higher decay) and higher contribution of recent data. We report on balanced accuracy, recall and Cum.S.P. (for visibility purposes Cum.S.P. was multiplied by $10^2$).

\begin{figure}[tp!]
 \centering
 \includegraphics[width=.75\textwidth]{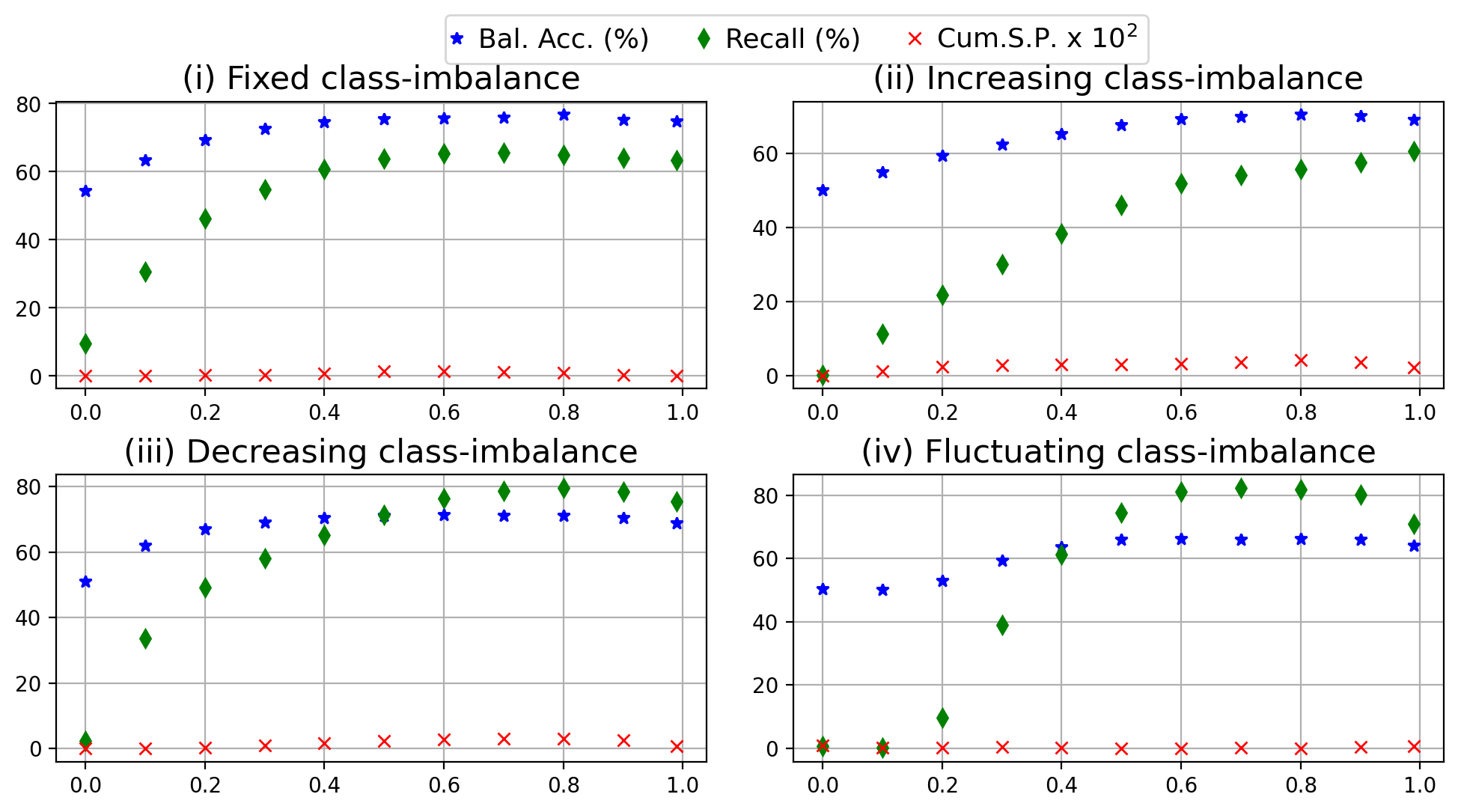}
 \caption{Impact of $\lambda$ on the synthetic datasets of varying class imbalance}
 \label{fig:synth_datasets_lamda_results}
\end{figure}

The results are displayed in Figure~\ref{fig:synth_datasets_lamda_results}, for each data stream. In the first case, where the class ratio is fixed over the stream (i) (25\%-75\%), low $\lambda$ values hurt the performance of \ours, while recall is less than 60\% for $\lambda < 0.4$. For $\lambda \geq 0.5$, balanced accuracy and recall are not affected significantly. In addition, for all values of $\lambda$ \ours's ability to mitigate unfair outcomes is not affected. For the increasing (ii) and decreasing (iii) class imbalance cases, we observe a steady increase of recall as $\lambda$ increases. Finally, for the fluctuating case (iv) of class imbalance, we denote that small values of $\lambda$ hurt the performance as in previous cases, but very high values also hurt the minority class. In the fluctuating case, the positive class is the minority in the beginning and afterwards changes into majority and then minority again. High $\lambda$ values allow the model to consider more history; therefore, the class imbalance weights which are assigned by \ours~are not adapted to recent changes.

\noindent\textbf{Conclusion:} We have seen that \ours~is able to handle varying class ratios over time and at the same time maintain low discriminatory outcomes. Values of $\lambda$ have a direct impact on \ours~performance. Small values of $\lambda$ generate fluctuating class imbalance weights which hurt the model's performance. Furthermore, in some cases values close to 1 are also not appropriate; therefore, we suggest users to consider values in the range of $[0.5, 0.9]$ which produce good predictive performance.


%% file: impact_of_M.tex
\begin{figure}[t!]
 \centering
 \includegraphics[width=.75\textwidth]{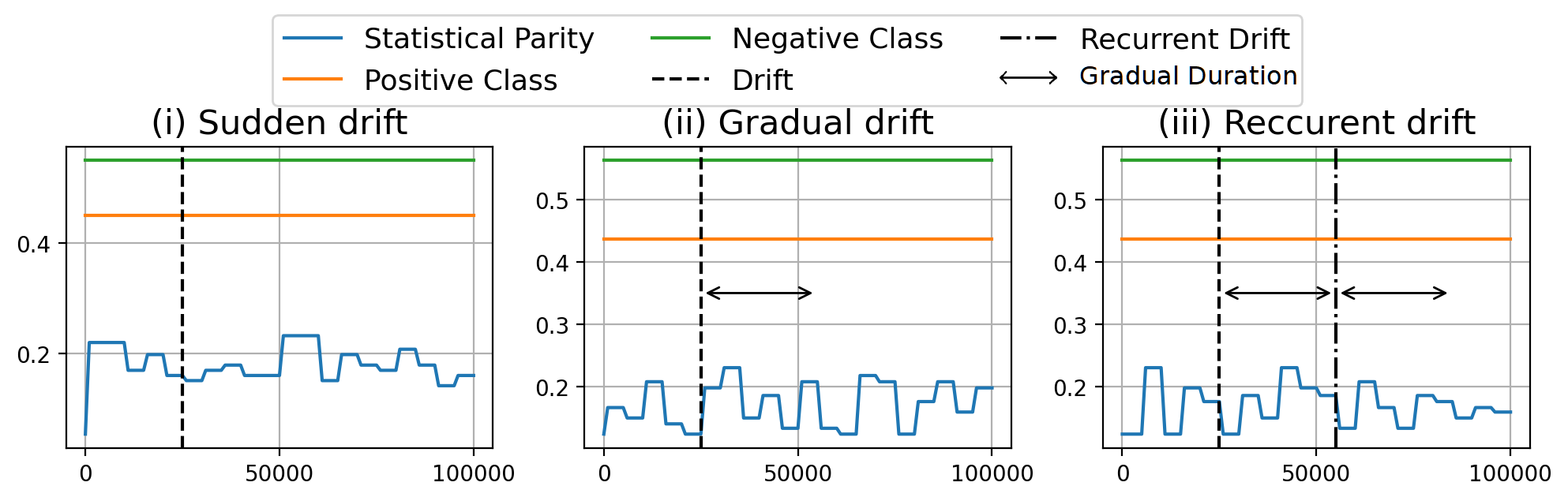}
 \caption{Synthetic datasets of various concept drifts (employed for $M$ evaluation)}
 \label{fig:synth_datasets_M}
\end{figure}

In the following section, we analyze \ours's ability to overcome various types of concept drifts and remain fair by taking into consideration the hyper-parameter $M$. $M$ is the size of sliding window which is used for adjusting the fair boundary. For this analysis, we experiment with 3 types of concept drifts: i) the sudden (abrupt) drift in which the mean values of the attributes of each class are shifted by a large number, ii) the gradual drift in which the mean values are shifted continuously by small increment and, iii) the recurrent drifts in which we perform gradual drift and after a while, we set the mean values back to their original values. The generated data streams are shown in Figure~\ref{fig:synth_datasets_M}. To make the evaluation more challenging, we fluctuate the encoded bias (statistical parity) in the datasets over time since $M$ is used for the fair adjustment of the boundary. For our analysis, we select various values of $M \in [500, 1000, 2000, 5000, 10000]$. For predictive performance we report on balanced accuracy and for fairness on Cum.S.P. 
\begin{figure}[t!]
 \centering
 \includegraphics[width=.75\textwidth]{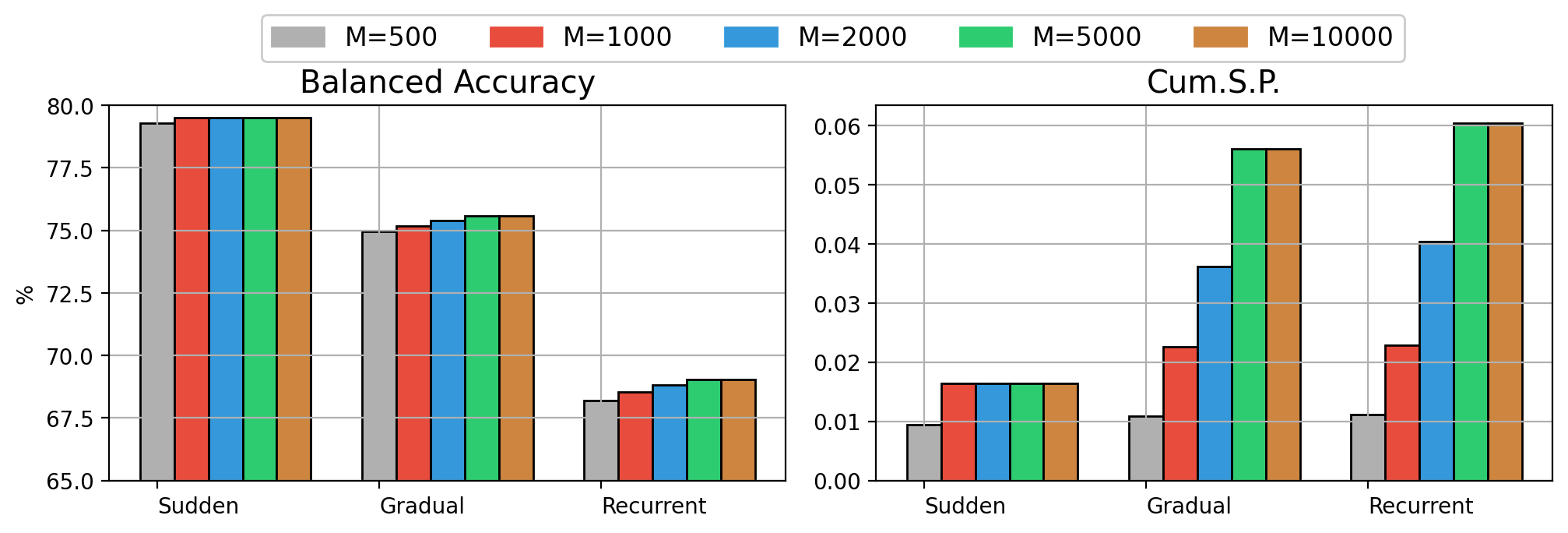}
 \caption{Impact of $M$ on the synthetic datasets of different types of concept drifts}
 \label{fig:perf_synth_datasets_M}
\end{figure}

The results of the experiments are shown in Figure~\ref{fig:perf_synth_datasets_M}. \ours's behaviour is similar across the different types of concept drifts i.e., large values of $M$ achieve better predictive performance in contrast to smaller values of $M$; nonetheless, discriminatory outcomes are affected significantly. 
In the sudden drift scenario, we observe that for $M \geq 1000$ the predictive scores and Cum.S.P. values do not change. This occurs while there is only one concept drift and the sliding windows ($M \geq 1000$) were not able to discard instances which appeared before the concept drift. The small window ($M = 500$) was able to accommodate new instances after the drift and discard the older ones. For the gradual and recurrent concept drifts, the behavior is similar to the sudden drift; however, the difference in terms of fairness is enlarged. Gradual drift is not an instant incident, which means that small $M$ values will be able to adjust the decision boundary faster. For $M=1000$ and $M=2000$ the windows were able to discard old instances but this did not happen for $M \geq 5000$. 

\noindent\textbf{Conclusion:} In this section, we observed how the sliding window size $M$ impacts the discriminatory behavior of \ours~under various concept drifts. We have seen that small values of $M$ were able to produce fairer outcomes while they were able to discard older information faster. Although the predictive performance among various values of $M$ was not significantly different, we believe this is due to the simplicity of the synthetic data. Small values of $M$ may produce fairer results but they will also deteriorate the predictive performance of the model. On the other hand, very large values of $M$, will make the model unable to adjust effectively the decision boundary in the presence of concept drifts.

%% file: conclusions.tex
In this paper, we proposed FABBOO, an online fairness-aware learner for data streams with class imbalance and concept drifts. Our approach changes the training distribution online taking into account class imbalance and tackles unfairness by adjusting the decision boundary on demand, based on different parity-based fairness notions. 
Our experiments exhibited that our approach outperforms other methods in a variety of datasets w.r.t. both predictive- and fairness-performance. In addition, we showed that recent fairness-aware stream learning methods reject the minority class at large to ensure fair results. On the contrary, our class imbalance-oriented approach effectively learns both classes and fulfills different fairness criteria while achieving good predictive performance for both classes. Furthermore, our cumulative definitions of fairness over the stream enable the model to mitigate long-term discriminatory effects, in contrast to a short-term definition like in CFBB and MS which are unable to deal with cumulative discrimination, discrimination propagation  and its reinforcement in the stream. Also, \ours~is able to maintain better predictive performance in contrast to OSBoost and CSMOTE in the presence of class imbalance.

We have provided a detailed analysis on \ours's hyper-parameter selection. \ours's ability to maintain fair outcomes is directly connected to the sliding window size $M$ as we have seen in Sec.~\ref{sec:impact_of_M}. Furthermore, the number of weak learners $N$ is not of great importance as long as it is sufficiently large (Sec.~\ref{sec:impact_of_t}). Finally, parameter $\lambda$  should not receive too low or too high values in the range of $[0, 1)$ as we have seen in Sec.~\ref{sec:impact_of_L}.  

As part of our future work, we plan to embed the decision boundary adjustment directly into the training phase by altering the weighted training distribution, as proposed in~\cite{iosifidis2019adafair}. Finally, although we assumed that the minority class was not fixed over the stream, we have assumed that the protected group is fixed over the stream. In our experiments \ours~was able to avoid reverse discrimination; however, we intend to waive this possibility by also considering adjusting the threshold on the other discriminated group if such incident occurs. 